\documentclass[journal]{IEEEtran}
\usepackage{alltt}
\usepackage{arydshln}
\usepackage{amsfonts}
\usepackage{amssymb}
\usepackage{amsthm}
\usepackage{algorithm}
\usepackage{algorithmic}

\usepackage{array}
\usepackage{bm}
\usepackage{booktabs}
\usepackage{cite}
\usepackage{color}
\usepackage{colortbl}
\usepackage{comment}
\usepackage[cmex10]{amsmath}
\usepackage{epstopdf}
\usepackage{enumerate}
\usepackage{flushend}
\usepackage{graphicx}
\usepackage{multicol}
\usepackage{multirow}
\usepackage{makecell}
\usepackage{overpic}
\usepackage{picinpar}
\usepackage{pifont}
\usepackage{pbox}
\usepackage{soul}
\usepackage{siunitx}
\usepackage{subfig}
\usepackage{tabularx}
\usepackage{tabularray}
\usepackage{url}
\usepackage{color}
\usepackage{xcolor}

\newtheorem{property}{Property}

\newtheorem{assumption}{Assumption}
\newtheorem{remark}{Remark}

\ifCLASSINFOpdf{}
\else
\fi

\usepackage{cite}

\makeatletter  
\let\NAT@parse\undefined{}
\makeatletter

\usepackage[colorlinks,linkcolor=blue,citecolor=blue,urlcolor=black]{hyperref}
\usepackage{cleveref}

\begin{document}
\title{
    RINGO:\ Real-time Navigation with a Guiding Trajectory for Aerial Manipulators in Unknown Environments
}

\author{
  \vskip 1em
  Zhaopeng Zhang, 
  Shizhen Wu,     
  Chenfeng Guo,  
  Yongchun Fang,  \emph{Senior Member,~IEEE}, \\
  Jianda Han, \emph{Member,~IEEE},
  Xiao Liang, \emph{Senior Member,~IEEE}
\thanks{\emph{Corresponding author: Xiao Liang}}

\thanks{The authors are with the Institute of Robotics and Automatic Information System, College of Artificial Intelligence, and Tianjin Key Laboratory of Intelligent Robotics, Nankai University, Tianjin 300350, China, and also with Institute of Intelligence Technology and Robotic Systems, Shenzhen Research Institute of Nankai University, Shenzhen 518083, China
(e-mail: \url{zhangzp@mail.nankai.edu.cn}; \url{szwu@mail.nankai.edu.cn}; \url{guocf@mail.nankai.edu.cn}; \url{liangx@nankai.edu.cn}; \url{fangyc@nankai.edu.cn}; \url{hanjianda@nankai.edu.cn})}
}

\maketitle
\IEEEpeerreviewmaketitle\
\begin{abstract}
  Motion planning for aerial manipulators in constrained environments has typically been limited to known environments or simplified to that of multi-rotors, which leads to poor adaptability and overly conservative trajectories.
  This paper presents RINGO:~Real-time Navigation with a Guiding Trajectory, a novel planning framework that enables aerial manipulators to navigate unknown environments in real time.
  The proposed method simultaneously considers the positions of both the multi-rotor and the end-effector.
  A pre-obtained multi-rotor trajectory serves as a guiding reference, allowing the end-effector to generate a smooth, collision-free, and workspace-compatible trajectory.
  Leveraging the convex hull property of B-spline curves, we theoretically guarantee that the trajectory remains within the reachable workspace.
  To the best of our knowledge, this is the first work that enables real-time navigation of aerial manipulators in unknown environments.
  The simulation and experimental results show the effectiveness of the proposed method. The proposed method generates less conservative trajectories than approaches that consider only the multi-rotor.
\end{abstract}

\begin{IEEEkeywords}
  Aerial manipulator, motion planning, trajectory optimization.
\end{IEEEkeywords}

\section{Introduction}\label{sec:introduction}
\IEEEPARstart{A}{erial} manipulators, typically consisting of a multi-rotor and a robotic arm, integrates the strengths of both components: the robotic arm provides the multi-rotor with manipulation capabilities, while the multi-rotor overcomes the fixed workspace limitation of the robotic arm, enhancing the aerial manipulator's flexibility for large-scale movements. 
In recent years, aerial manipulators have attracted considerable attention for the robust control~\cite{zhang2019tie,wang2023tro}, visual servo~\cite{zhong2020tie,chen2024tie}, and various practical applications~\cite{lee2020icra,gonzalez2024tmech,bodie2021tro}.

Motion planning is a fundamental problem in robotics, aiming to generate collision-free and dynamically feasible trajectories for robotic systems. Although some studies have addressed the motion planning problem for aerial manipulators, there are still two key issues that need to be addressed. Firstly, the motion planning algorithm for aerial manipulators is typically conducted in known environments~\cite{kim2019sample,seo2017local,cao2024motion,caballero2018first,deng2025arxiv,zhang2025tmech}. However, addressing the motion planning problem in an unknown environment presents significant challenges in terms of computational complexity and real-time performance.
Secondly, the motion planning for an aerial manipulator in constrained environments is, in some cases, simplified to that of a multi-rotor~\cite{cao2024motion}. However, this approach, which encloses the entire system within a large bounding sphere, tends to yield overly conservative trajectories.
While minimizing the sphere's radius by retracting the robotic arm can mitigate this issue, it introduces a periodic planning approach, potentially increasing the time required for the arm to reach its goal state~\cite{ollero2021tro}.
\begin{figure}[t]
  \centering
  \subfloat[\footnotesize aerial manipulator passing through a ring-shaped obstacle]{
    \includegraphics[width=0.235\textwidth]{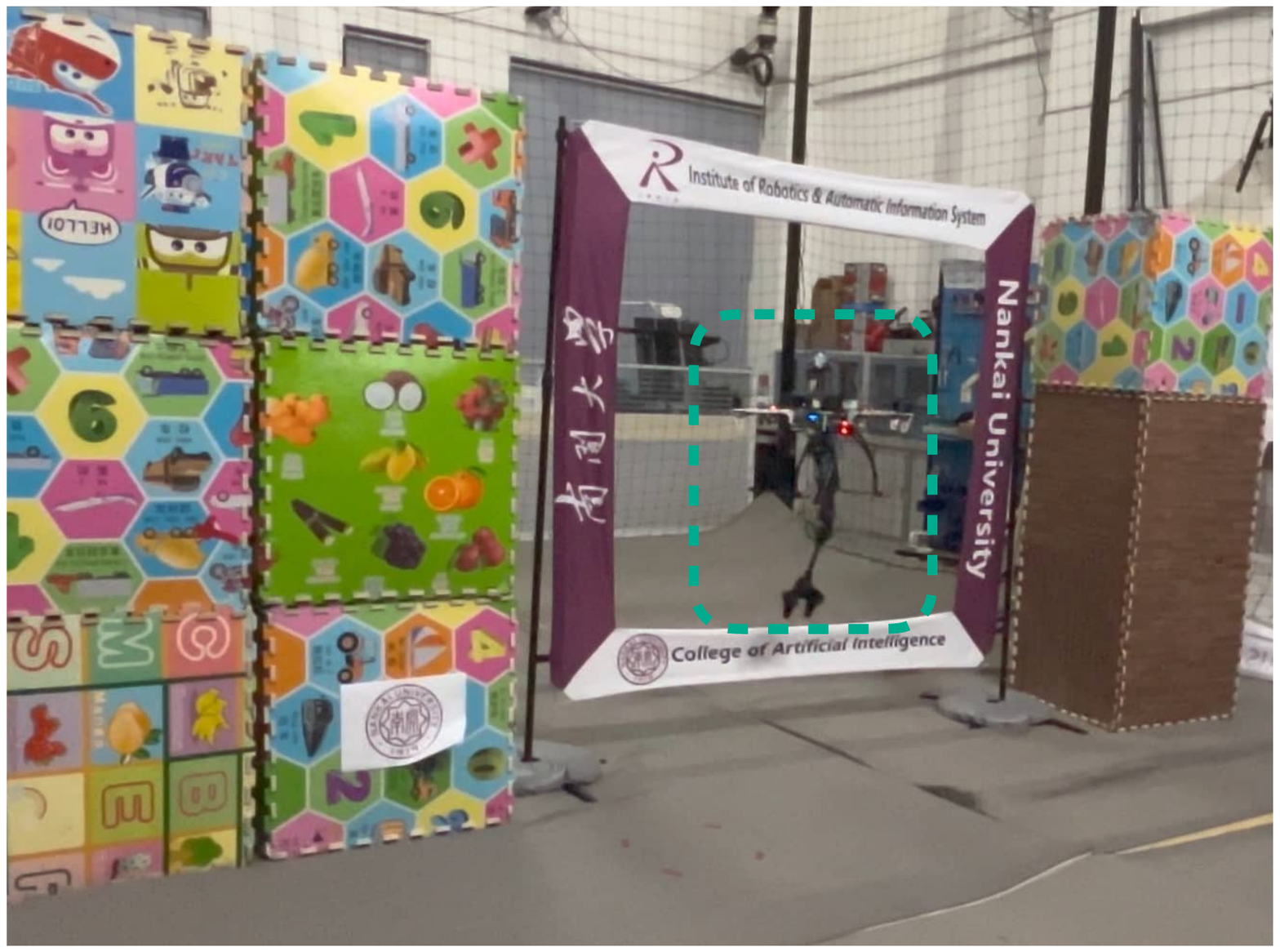}\label{fig:intro_exp}
  }
  \hspace{-0.4cm}
  \subfloat[\footnotesize visualization in RViz]{
      \includegraphics[width=0.235\textwidth]{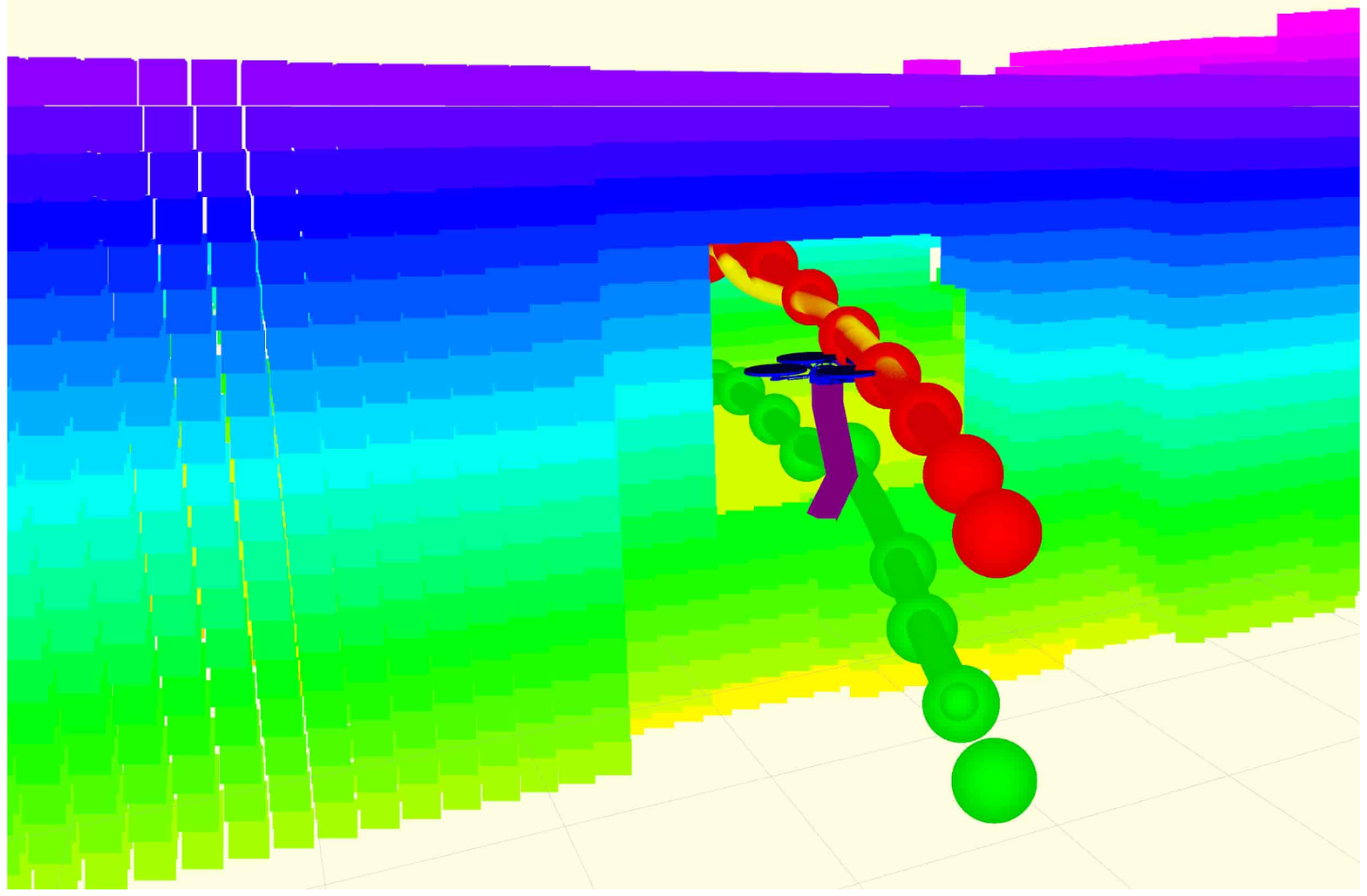}\label{fig:intro_rviz}
  }
  \caption{Our proposed method is validated in a real-world environment. Experimental details are given in Sec.\ \ref{sec:implementation_results}.}\label{fig:intro_exp_rviz}
\end{figure}

Building on the extensive previous research and the vibrant open-source community in the field of multi-rotor motion planning, we propose a novel motion planning method, called \textbf{R}eal-t\textbf{I}me \textbf{N}avigation with a \textbf{G}uiding traject\textbf{O}ry (\textbf{RINGO}) for aerial manipulators in unknown environments.
Our proposed method employs a leader-follower-inspired motion planning framework for the aerial manipulator. Based on a previously planned and parameterized B-spline trajectory of the multi-rotor, we then plan the trajectory for the end-effector. 
Firstly, the initial trajectory for the end-effector is generated by a second-order Bézier curve from the initial position to the goal position and refined into a B-spline curve.
Then, the gradient-based optimization method is used to optimize the trajectory of the end-effector. The linear and convex-hull properties of the B-spline curve guarantee that the trajectory is smooth, collision-free, and compatible with the available workspace.
Additionally, considering the motion capabilities related to the yaw angle of the multi-rotor, we incorporate an extra yaw rate cost into the trajectory optimization problem.

Compared with the existing works, our proposed method is able to generate the trajectory for the aerial manipulator in real-time without reducing the system to a multi-rotor-only model. 
Through theoretical analysis, the trajectory generated by the proposed method is guaranteed to be within the workspace and collision-free.
The simulation and experimental results show that the proposed method can generate a collision-free and workspace-compatible trajectory for the aerial manipulator in real time.
The main contributions of this paper are listed as follows:
\begin{enumerate}
\item Unlike most existing works that plan trajectories for aerial manipulators in known constrained environments, this paper presents the first real-time motion planning algorithm for aerial manipulators in unknown environments without degrading the problem to that of a multi-rotor.
\item With the convex hull property and the linear property of the B-spline curve, the trajectory generated by the proposed method can be guaranteed to be collision-free and workspace-compatible theoretically. 
\end{enumerate}

The rest of the paper is organized as follows: 
Section~\ref{sec:relatedwork} presents the related work on motion planning for both multi-rotors and aerial manipulators. 
Section~\ref{sec:preliminary} introduces some preliminaries including the aerial manipulator system and B-spline curve.
Section~\ref{sec:algorithm_overview} and~\ref{sec:trajectory_optimization} present the main part of the proposed method. 
In Section~\ref{sec:implementation_results}, the details of the algorithm implementation, simulation and experimental results are presented. 
Finally, Section~\ref{sec:conclusion} concludes the paper and outlines future directions.


\section{Related Work}\label{sec:relatedwork}
\subsection{Motion Planning for Multi-rotors}
By invoking the differential flatness property\cite{mellinger2011minsnap}, the motion planning problem is simplified to consider only the position and the yaw angle of the multi-rotor\cite{zhou2021raptor}. 
In cases where a 360-degree LiDAR, rather than a front-view camera, is mounted on the multi-rotor\cite{zhou2019robust}, the yaw angle of the multi-rotor can be disregarded. Consequently, the motion planning problem is further simplified to find a collision-free and dynamically feasible trajectory for a point in $\mathbb{R}^3$, with obstacle avoidance ensured by inflating the obstacles.
Alternatively, the multi-rotor can be enclosed within an ellipsoid\cite{liu2018search} or a convex polyhedron\cite{han2021racing} to reduce the conservatism of the trajectory by formulating the problem in $SE(3)$.
Most existing studies on motion planning can be divided into two main stages: path search and trajectory optimization. 
In\ \cite{zhou2019robust}, the initial path is generated using the hybrid A\textsuperscript{*} algorithm, followed by trajectory optimization through a gradient-based method after the trajectory has been parameterized as a B-spline curve.
Similarly, the jump point search (JPS) algorithm is employed to determine the initial waypoints for the multi-rotor, and the trajectory is optimized by parameterizing it as a Bézier curve, as demonstrated in\ \cite{tordesillas2021faster}.

\subsection{Motion Planning for Aerial Manipulators}
Motion planning for aerial manipulators involves generating safe and feasible trajectories that account for both the multi-rotor and the robotic arm. 
Some studies adopt decoupled motion planning frameworks, in which the multi-rotor and the manipulator are planned in separate stages~\cite{ollero2021tro}.
For instance, Kim \textit{et al.}~\cite{kim2019sample} integrate informed-RRT\textsuperscript{*} with the local planner in~\cite{seo2017local} to generate collision-free trajectories in known environments. 
\text{Cao} \textit{et al.}~\cite{cao2024motion} propose a two-stage decoupled method for pick-and-place tasks, where the aerial manipulator is enclosed within a large sphere to simplify collision avoidance.

More recent efforts have incorporated whole-body planning strategies. Alvaro \textit{et al.}~\cite{caballero2018first} consider the kinematic model of the aerial robotic system with two arms for long-reach manipulation and use the RRT\textsuperscript{*}-based method to plan the trajectory for the multi-rotor and the robotic arm in a known environment.
\text{Deng} \textit{et al.}~\cite{deng2025arxiv} propose a dynamic ellipsoidal approximation method that adapts to varying manipulator configurations, enabling precise collision checking for the aerial manipulator with a delta arm. However, this method may not generalize well to serial-link arms due to their asymmetric and configuration-dependent collision volumes.
Zhang \textit{et al.}~\cite{zhang2025tmech} formulated a coupled motion planning method by enclosing the aerial manipulator within a convex polyhedron and optimizing its trajectory in structured known environments. 

Meanwhile, other works\cite{tognon2018control,luo2023timeoptimal,jakewelde2020coordinate,jakewelde2021dynamic} explore task-constrained planning without addressing collision avoidance, limiting their applicability in real-world environments.


\section{Preliminary}\label{sec:preliminary}
\subsection{Aerial Manipulator}
\begin{figure}[htbp]
  \centering
  \subfloat[\footnotesize three coordinate frames]{
    \includegraphics[width=0.22\textwidth, trim=0 0 0 0, clip]{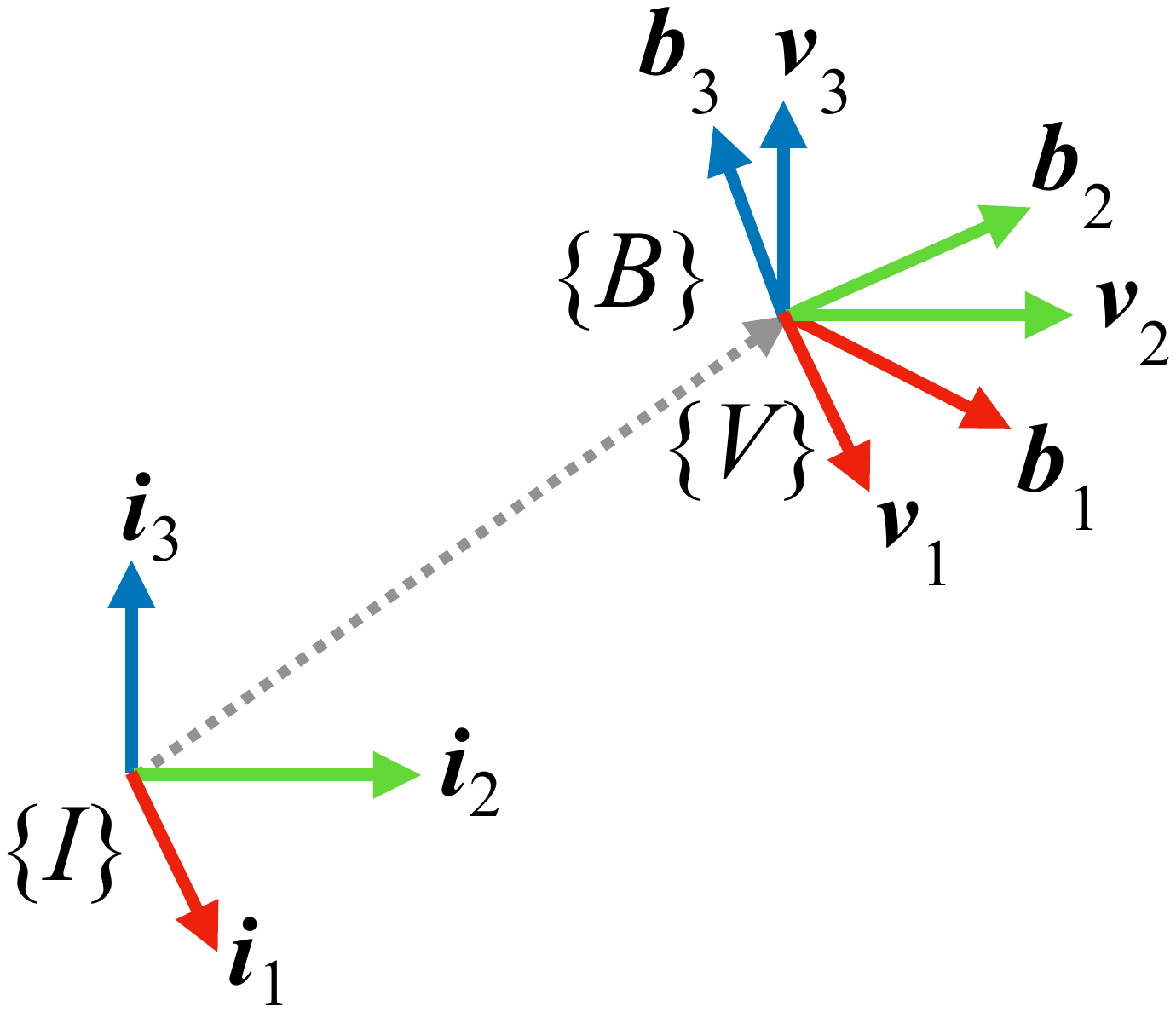}\label{fig:illustration_frames}
  }
  \hspace{-0.5cm}
  \subfloat[\footnotesize illustration of the Assumption\ \ref{assumption:b3_e3}]{
    \includegraphics[width=0.24\textwidth, trim=0 0 0 0, clip]{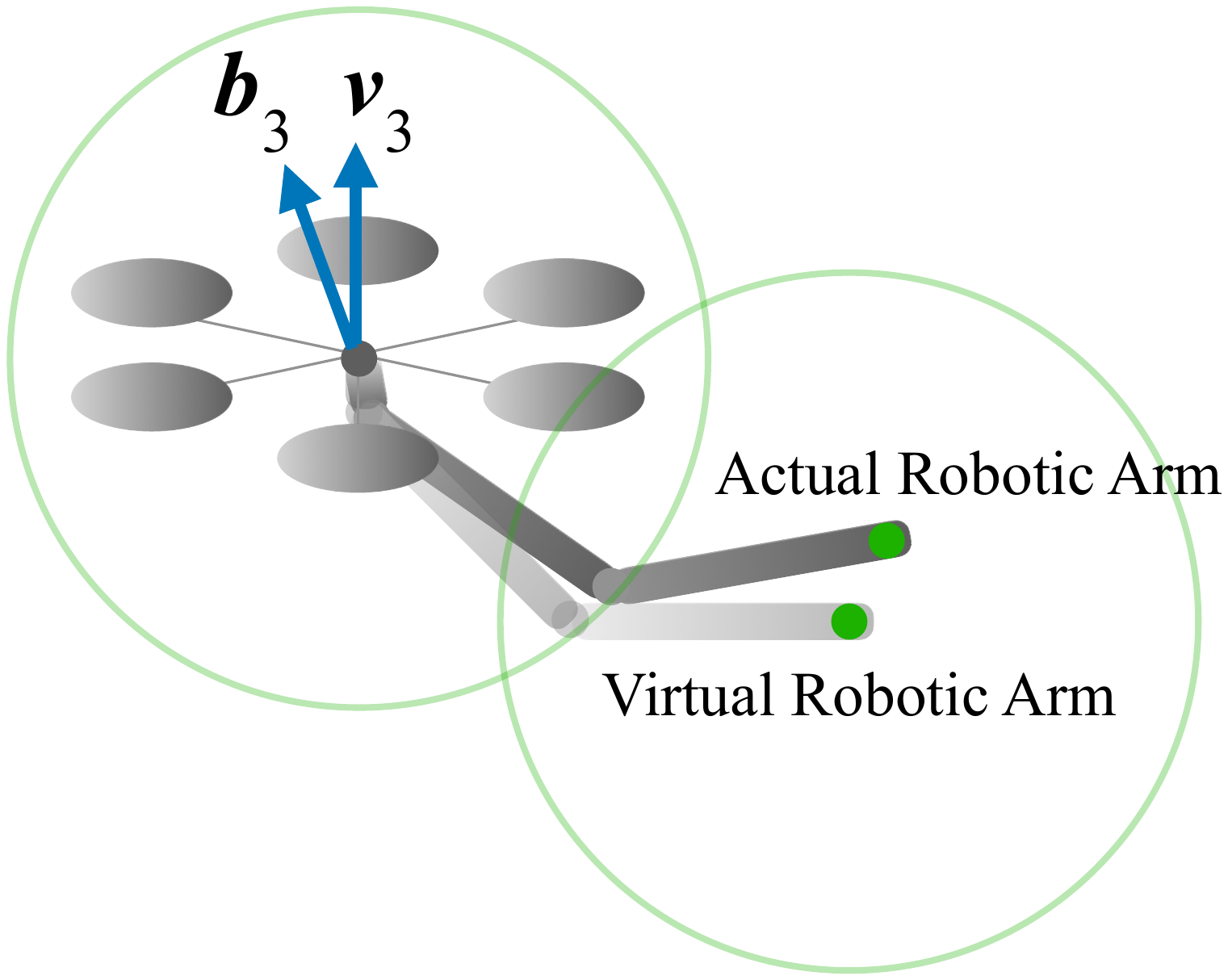}\label{fig:assumption_b3_e3}
  }
  \caption{Illustration of the aerial manipulator and the defined coordinate frames.}\label{fig:illustration_of_preliminary}
\end{figure}

In this paper, three frames are defined as $\{I\}=\{ \bm i_1, \bm i_2, \bm i_3 \}$, $\{B\}=\{ \bm b_1, \bm b_2, \bm b_3 \}$ and $\{V\}=\{ \bm v_1, \bm v_2, \bm v_3 \}$ which represent the inertia frame, the body-fixed frame, and the virtual body-fixed frame, respectively, which is presented in Fig.\ \ref{fig:illustration_frames}. The three axes of coordinate frame $\{V\}$ are parallel to the corresponding three axes of coordinate frame $\{I\}$ and the origin of frame $\{V\}$ coincides with that of frame $\{B\}$.

The aerial manipulator combines a multi-rotor and a 2-DoF pitch-pitch robotic arm, as shown in Fig.\ \ref{fig:assumption_b3_e3}, whose degree of freedom is 8. The state variables of the aerial manipulator are defined as follows:
\begin{align}
  \bm q = \left\{\begin{aligned} \bm x \\ R \\ \bm \theta \end{aligned}\right\} \in \mathbb{R}^3 \times SO(3) \times \Theta, 
\end{align}
where $\bm x \in \mathbb{R}^3$ is the multi-rotor's position with respect to the inertial frame $\{I\}$, $R \in SO(3)$ is the rotation matrix from the body-fixed frame $\{B\}$ to the inertial frame $\{I\}$, $\bm \theta = \{\theta_1, \theta_2\} \in \Theta$ represents the joint angles of the robotic arm and $\Theta = \Theta_1 \times \Theta_2$. $\theta_i$ represents the joint angle of the $i$-th joint, where $\Theta_i \subset \mathbb{R} (i = 1, 2)$ specifies the allowable range of $\theta_i$.
$\bm x_e \in \mathbb{R}^3$ is the position of the end-effector with respect to the inertial frame $\{I\}$, which is defined as follows:
\begin{align}
  \bm x_e = \bm x + R (\bm b_3, \psi) {}^b \bm x_e(\bm \theta).\label{eq:xe_before_assumption}
\end{align}
The rotation matrix $R(\bm b_3, \psi)$ can be divided into two parts, including the tilt motion and the yaw angle $\psi$, as follows\cite{Watterson2020ISRR}:
\begin{align*}
  R (\bm b_3, \psi) = H_2(\bm b_3) H_1(\psi).
\end{align*}

{\assumption\ The direction of the thrust $\bm b_3$ is not too far from $\bm v_3$ of the virtual body-fixed frame.\ \label{assumption:b3_e3}}

With the aforementioned assumption, it can be derived that
\begin{align*}
  R (\bm b_3, \psi) \approx R (\bm v_3, \psi) = H_2(\bm v_3) H_1(\psi) = H_1(\psi).
\end{align*}
The position of the end-effector $\bm x_e$ in\ \eqref{eq:xe_before_assumption} can be derived as
\begin{align}
  \bm x_e \approx \bm x + R (\psi) {}^b \bm x_e(\bm \theta)
          = \bm x + {}^v \bm x_{e}(\bm \theta^+),\label{eq:xe_after_assumption}
\end{align}
where $\bm \theta^+ = \{\psi, \bm \theta\}{}$ is defined as the generalized joint angle vector. ${}^v \bm x_e \in \mathbb{R}^3$ represents the end-effector position with respect to the virtual body-fixed frame $\{V\}$, which is denoted as $\bm x_{ve}$ in the following context.

As shown in Fig.\ \ref{fig:assumption_b3_e3}, the light-colored robotic arm represents the virtual arm, while the dark-colored one corresponds to the actual arm. The two green dots indicate the positions of the end-effectors associated with the virtual and actual arms, respectively. Notably, two appropriately sized spheres centered at the positions of the multi-rotor and the end-effector are sufficient to enclose the entire aerial manipulator system. This geometric abstraction facilitates subsequent safety margin analysis and collision checking in the planning process.

\subsection{B-spline Curve}\label{subsec:bspline_curve}
\begin{figure}[htbp]
  \centering
  \begin{overpic}[width=0.48\textwidth, trim=0 5cm 0 3cm, clip]{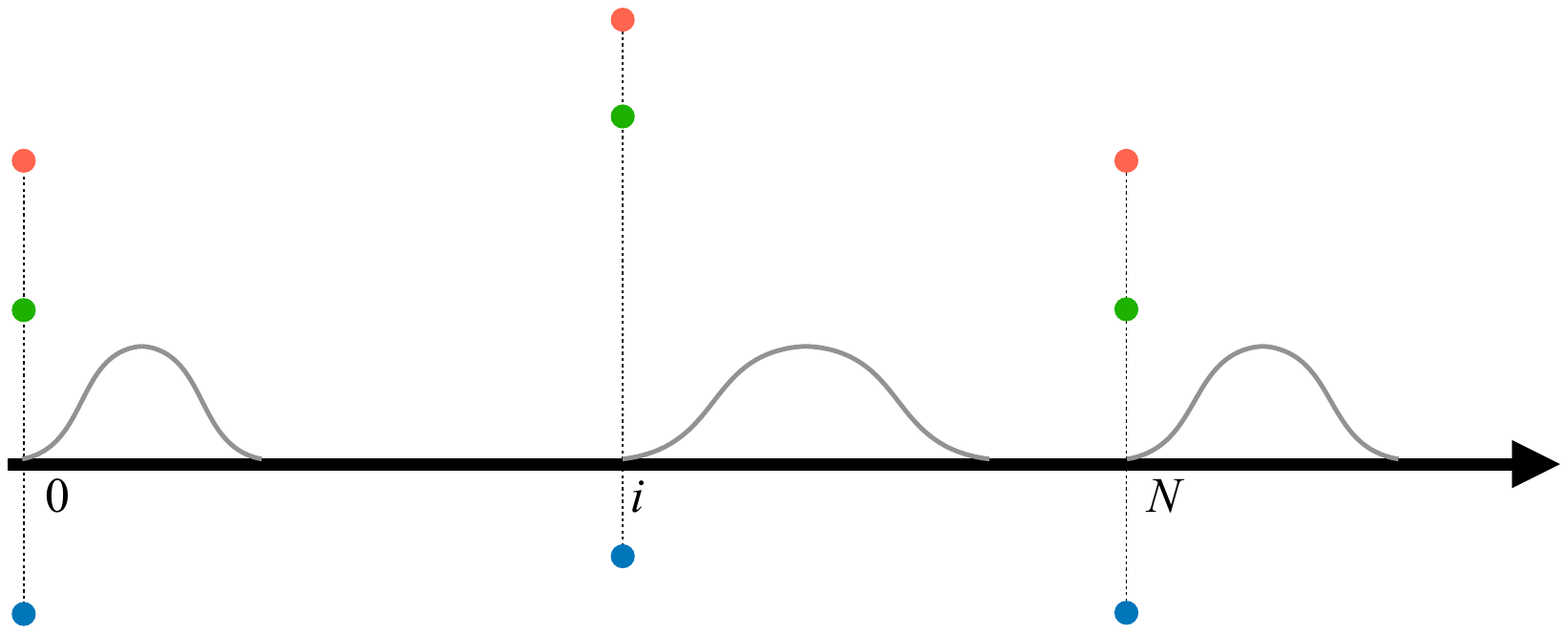}
    \put(7, 35){$\mathcal{X}_0$}
    \put(7, 25){$\mathcal{Q}_0$}
    \put(7, 5){$\mathcal{E}_0$}
    \put(42, 42){$\mathcal{X}_i$}
    \put(42, 35){$\mathcal{Q}_i$}
    \put(42, 8){$\mathcal{E}_i$}
    \put(73, 35){$\mathcal{X}_N$}
    \put(73, 25){$\mathcal{Q}_N$}
    \put(73, 5){$\mathcal{E}_N$}
    \put(95, 18){$t$}
  \end{overpic}
  \caption{Illustration of the linear property of the B-spline curve. }\label{fig:bspline_linear_property}
\end{figure}
An $s$-order B-spline curve is determined by a set of $N + 1$ control points $\mathcal{E}=\{\mathcal{E}_0, \mathcal{E}_1, \ldots, \mathcal{E}_N\}$ and a knot vector $\mathcal{T} = [t_0, t_1, \ldots, t_{M}]{}^\top \in \mathbb{R}^{M+1}_+$, where $\mathcal{E}_i \in \mathbb{R}^3$, $t_i \in \mathbb{R}_+$ and $M = N + s + 1$.

{\property\ If two B-spline curves with the same order share the same knot vector, the linear combination of them is still a B-spline curve with the same order.\label{property:linear_combination}}

It is assumed that the trajectory of the multi-rotor's position $\bm x(t)$ is a B-spline curve, determined by one set of control points $\mathcal{X}=\{\mathcal{X}_0, \mathcal{X}_1, \ldots, \mathcal{X}_N\}$, and the trajectory of the end-effector's position $\bm x_e(t)$ is also a B-spline curve, determined by another set of control points $\mathcal{Q}=\{\mathcal{Q}_0, \mathcal{Q}_1, \ldots, \mathcal{Q}_N\}$. The two B-spline curves share the same time knots vector $\mathcal{T}$ and are defined as
\begin{align}
    \bm x(t) &= \sum_{i=0}^{N} \bm B_i(t) \mathcal{X}_{i}, \\
  \bm x_e(t) &= \sum_{i=0}^{N} \bm B_i(t) \mathcal{Q}_{i},
\end{align}
where $\bm B_i(t)$ is the B-spline basis function of order $s$. Then, $\bm x_{ve}(t)$ in\ \eqref{eq:xe_after_assumption} can be derived as
\begin{align}
  \bm x_{ve}(t) &= \bm x_e(t) - \bm x(t) \notag \\
                &= \sum_{i=0}^{N} \bm B_i(t) (\mathcal{Q}_i - \mathcal{X}_i ) 
                 = \sum_{i=0}^{N} \bm B_i(t) \mathcal{E}_{i},
\end{align}
where it can be concluded that $\bm x_{ve}(t)$ is also a B-spline curve with the control points $\mathcal{E}=\{\mathcal{E}_0, \mathcal{E}_1, \ldots, \mathcal{E}_N\}$ and share the same time knots vector $\mathcal{T}$ with $\bm x(t)$ and $\bm x_e(t)$, as shown in Fig.\ \ref{fig:bspline_linear_property}.


\section{Method Overview}\label{sec:algorithm_overview}
\subsection{Problem Statement}
The problem this paper aims to address is \textit{how to plan a collision-free trajectory for the aerial manipulator in real time within a constrained environment, given its initial and goal states}.

By inflating the obstacles, the collision avoidance constraint is formulated as two points in the $\mathbb{R}^3 \times \mathbb{R}^3$ space, where it is represented by the multi-rotor's position $\bm x$ and the end-effector position $\bm x_e$. As illustrated in Fig.\ \ref{fig:assumption_b3_e3}, two green circles are placed in the multi-rotor and the virtual end-effector. 
Once Assumption\ \ref{assumption:b3_e3} holds, $\bm b_3$ will remain sufficiently close to $\bm e_3$, and the large sphere attached to the virtual end-effector position will be adequately sized to encompass the actual robotic arm. As a result, the end-effector's trajectory will be decoupled from the multi-rotor's tilt motion.

\subsection{Main Method}
\begin{figure}[htbp]
  \centering
  \includegraphics[width=0.35\textwidth, trim=0 5cm 0 0, clip]{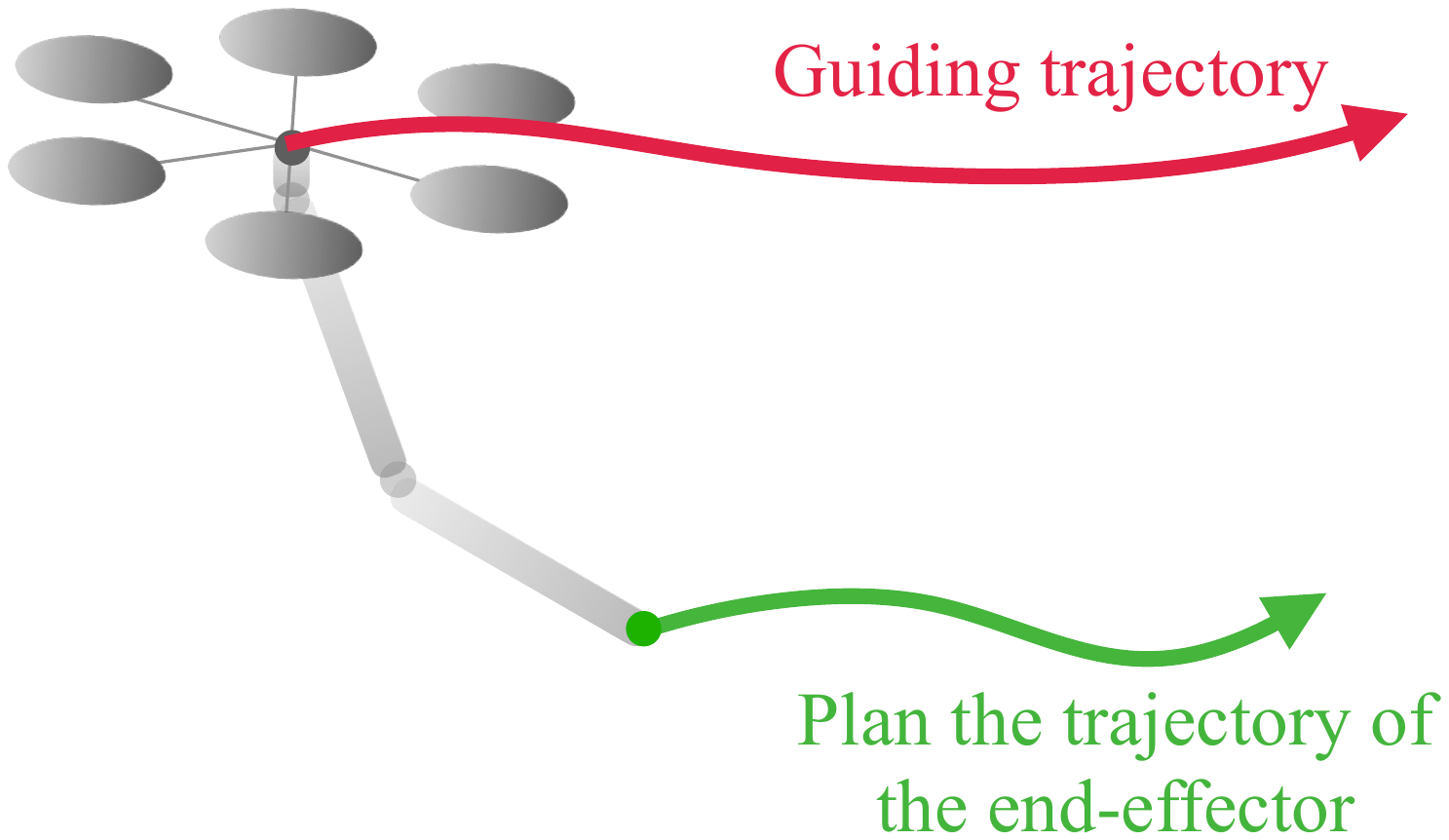}
  \caption{Illustration of the proposed guiding trajectory-based motion planning method.}\label{fig:guiding-plan}
\end{figure}
As shown in Fig.\ \ref{fig:guiding-plan}, the multi-rotor trajectory is firstly planned as the guiding trajectory and then parameterized as a B-spline curve, determined by the control points $\mathcal{X}$ and time knots $\mathcal{T}$. Then, the motion planning problem for the aerial manipulator is formulated as follows:
\begin{algorithm}[!htbp]
  \caption{Trajectory Planning for the robotic arm}\label{alg:traj_opt}
  \begin{algorithmic}[1]
    \REQUIRE{} the control points $\mathcal{X}$ and time knots $\mathcal{T}$ of the multi-rotor's trajectory $\bm x(t)$; the initial and goal position of the end-effector $\bm x_{e}(t_0)$, $\bm x_{e}(t_M)$.
    \ENSURE{} the trajectory of the end-effector $\bm x_{e}(t)$.
    \STATE{} \textbf{Generate the initial trajectory} for the end-effector by combining $\bm x(t)$ and a second-order Bézier curve detailed in Section\ \ref{subsec:initial_traj}.
    \STATE{} \textbf{Parameterize the trajectory} as a B-spline using control points $\mathcal{Q}$.
    \STATE{} \textbf{Optimize the control points $\mathcal{Q}$} with the gradient-based method detailed in Section\ \ref{sec:trajectory_optimization}.
    \STATE{} \textbf{Inverse Kinematics} to obtain the generalized joint trajectory $\bm \theta^+(t)$.
  \end{algorithmic}
\end{algorithm}

\subsection{Initial Trajectory Generation}\label{subsec:initial_traj}
\begin{figure}[htbp]
  \centering
  \includegraphics[width=0.35\textwidth, trim=0 0 0 4cm, clip]{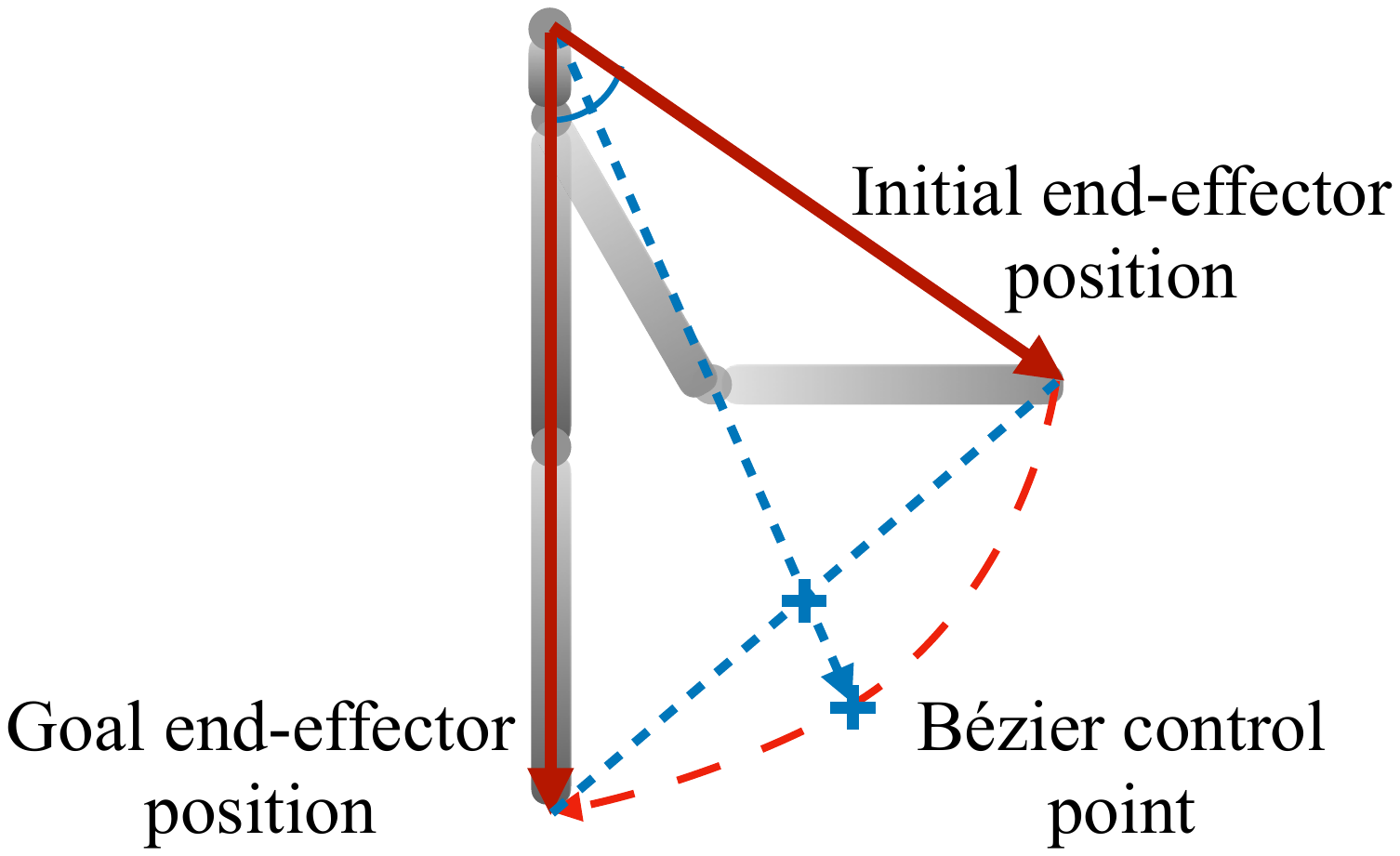}
  \caption{Initial trajectory for the end-effector.}\label{fig:initial_xve_traj}
\end{figure}

The initial trajectory of the end-effector $\bm x_{ve}(t)$ is generated by a second-order Bézier curve. The Bézier control point is determined by the initial and goal end-effector position with respect to the virtual body frame $\{V\}$, which is shown in Fig.\ \ref{fig:initial_xve_traj}. The Bézier control point is calculated as follows:
\begin{align}
  \bm P = \frac{1}{2} \lambda \left(\bm x_{ve}(t_0) + \bm x_{ve}(t_M)\right),\label{eq:bezier_control_point}
\end{align}
where $\lambda = \log \left(\frac{1}{2} \left| \arccos \left( \bm x^\top_{ve}(t_0) \bm x_{ve}(t_M) \right) + 1 \right| + 1 \right)$. 
It is equivalent to push the middle point of the initial and goal end-effector position far away. The bigger the angle between the initial and goal end-effector position vector, the larger the coefficient $\lambda$, which is designed to get the smooth trajectory in the joint space. 

The initial trajectory of the end-effector can be generated by combining the pre-obtained trajectory $\bm x(t)$ of the multi-rotor with the second-order Bézier curve for the end-effector. Subsequently, by refining the initial trajectory, the end-effector's trajectory is parameterized using B-spline control points $\mathcal{Q} = \{ \mathcal{Q}_0, \mathcal{Q}_1, \ldots, \mathcal{Q}_N \}$ along with time knots $\mathcal{T} = [t_0, t_1, \ldots, t_{M}]{}^\top$.


\section{Trajectory Optimization}\label{sec:trajectory_optimization}
According to the procedure in Section\ \ref{subsec:initial_traj}, the initial trajectory of the robotic arm with respect to $\{V\}$ is generated by a second-order Bézier curve. Then, the initial trajectory of the end-effector with respect to $\{I\}$ is obtained. It is also a B-spline curve sharing the same time knot vector $\mathcal{T}$ with the multi-rotor's trajectory $\bm x(t)$, which is parameterized using the control points $\mathcal{X}=\{\mathcal{X}_0, \mathcal{X}_1, \ldots, \mathcal{X}_N\}$. The obstacle avoidance problem is not considered in the initial trajectory generation. Then, the following trajectory optimization problem is formulated:
\begin{align}
    \min_{\mathcal{Q}} f = \lambda_s f_{s} + \lambda_w f_{w} + \lambda_y f_{y} + \lambda_d f_{d},\label{eq:opt_problem}
\end{align}
where $f$ represents the total cost, while $f_{s}$, $f_{w}$, $f_{y}$, and $f_{d}$ correspond to the smoothness cost, workspace cost, yaw rate cost, and obstacle avoidance cost, respectively. The weights for these cost components are denoted as $\lambda_s$, $\lambda_w$, $\lambda_y$, and $\lambda_d$. 
In the following context, both $\mathcal{Q} = \{\mathcal{Q}_0, \mathcal{Q}_1, \ldots, \mathcal{Q}_N\}$ and $\mathcal{E} = \{\mathcal{E}_0, \mathcal{E}_1, \ldots, \mathcal{E}_N\}$ will be utilized, and the relationship of them has been given in Section\ \ref{subsec:bspline_curve}.

\subsection{Workspace Cost}
\begin{figure}[htbp]
  \centering
  \includegraphics[width=0.36\textwidth]{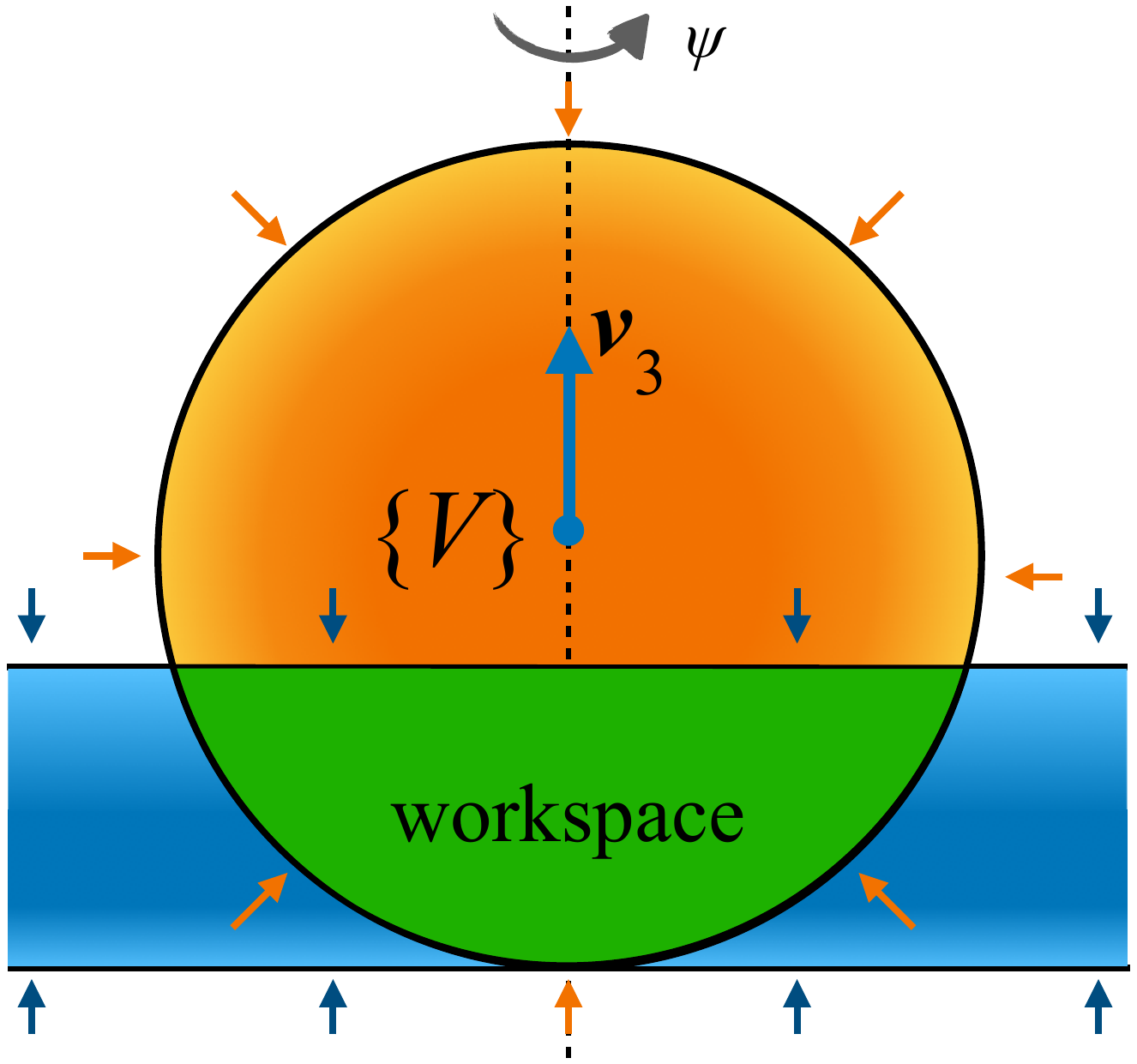}
  \caption{Illustration of the workspace. }\label{fig:workspace}
\end{figure}
The workspace for the end-effector $\bm x_{ve}(t)$, as depicted in Fig.\ \ref{fig:workspace}, is formulated as the intersection of a large sphere and two half-spaces, each defined by a plane, as the green region shown in the figure. The workspace, denoted as $\mathbb{W} \subset \mathbb{R}^3$, is a convex space. 

\begin{figure}[htbp]
  \centering
  \begin{overpic}[width=0.36\textwidth, trim=0 0 0 9cm, clip]{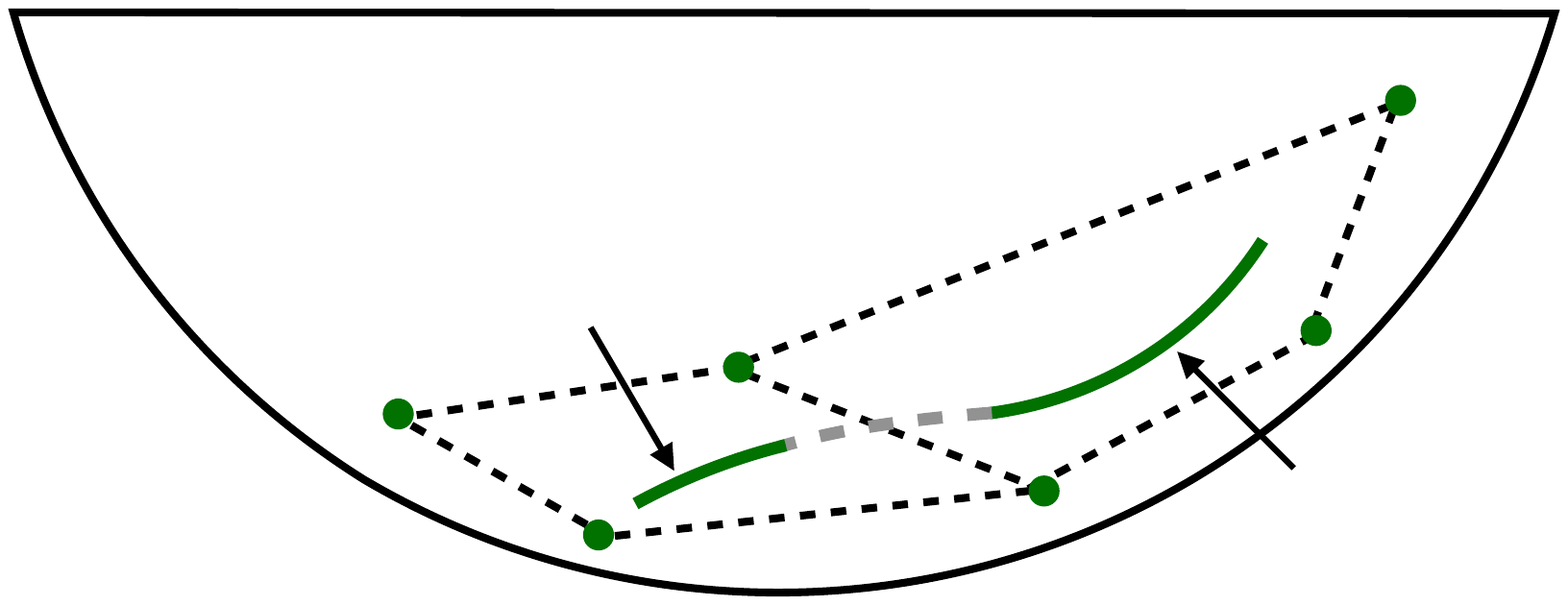}
    \put(80, 7){$\bm x_{ve1}(t)$}
    \put(27, 22){$\bm x_{ve3}(t)$}
    \put(88, 31){$\mathcal{E}_1$}
    \put(82, 15){$\mathcal{E}_2$}
    \put(68, 7){$\mathcal{E}_3$}
    \put(45, 20){$\mathcal{E}_4$}
    \put(36, 0){$\mathcal{E}_5$}
    \put(20, 16){$\mathcal{E}_6$}
  \end{overpic}
  \caption{Illustration of the convex hull in the workspace.}\label{fig:workspace_convexhull}
\end{figure}
\begin{remark}
  The shape of workspace is determined by the configuration of the robotic arm. For other configurations, the workspace can be formulated into different shapes as long as they are convex.
\end{remark}
Two segments of the B-spline curve $\bm x_{ve1}(t)$ and $\bm x_{ve3}(t)$ are shown in Fig.\ \ref{fig:workspace_convexhull}, which are determined by the control points set $\{\mathcal{E}_1, \mathcal{E}_2, \mathcal{E}_3, \mathcal{E}_4\}$ and $\{\mathcal{E}_3, \mathcal{E}_4, \mathcal{E}_5, \mathcal{E}_6\}$, respectively.

For the sake of simplicity, let $\mathbb{C}_1 \subset \mathbb{R}^3$ represent the interior space of a tetrahedron with $\{\mathcal{E}_1, \mathcal{E}_2, \mathcal{E}_3, \mathcal{E}_4\}$ as the vertices. Due to the convex hull property of the B-spline curve, the segment $\bm x_{ve1}(t) \in \mathbb{C}_1 \subset \mathbb{W}, t \in [t_1, t_2]$. Similarly, the whole B-spline curve $\bm x_{ve}(t)$, in which $t \in [t_0, t_M]$ is theoretically guaranteed to be in the convex workspace $\mathbb{W}$ theoretically, if all the control points are in the convex workspace.

According to the above analysis, for the purpose of constraining $\bm x_{ve}$ in the workspace, the corresponding cost is formulated as follows:
\begin{align}
  f_{w} = \sum_{i=s}^{N-s} \frac{1}{k} \log \left(e^{h_{o}k F_{o}(\mathcal{Q}_{i})} + e^{h_{l} k F_{l}(\mathcal{Q}_{i})} \right),\label{eq:workspace_cost}
\end{align}
where $F_{o}(\mathcal{Q}_{i})$ and $F_{l}(\mathcal{Q}_{i})$ are the approximated signed distance functions determined by the circle and the two lines, respectively. $k$ is a positive constant.
\begin{figure}[htbp]
  \centering
  \subfloat[\footnotesize variables of the functions]{
    \begin{overpic}[width=0.24 \textwidth]{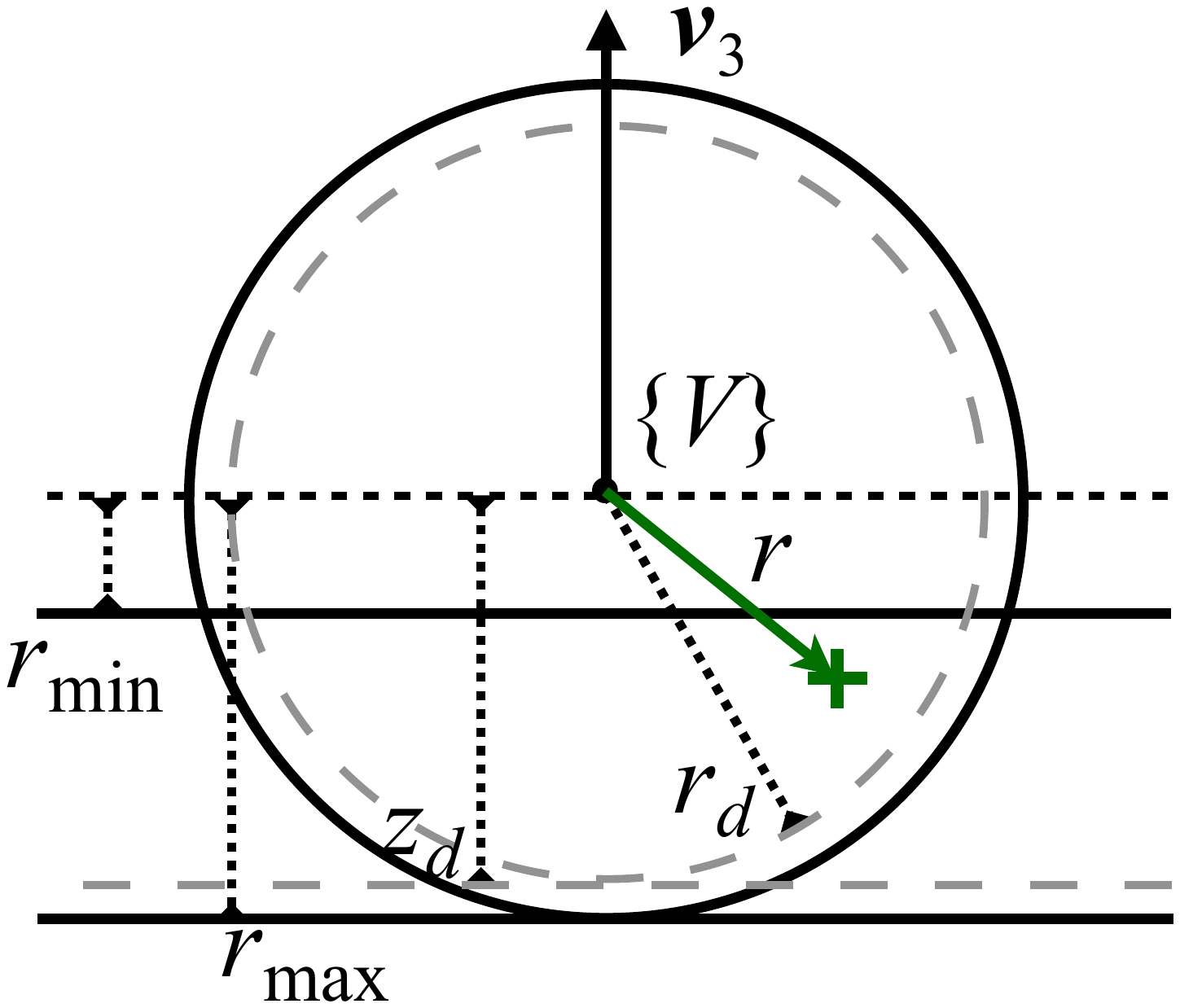}
      \put(64, 16){\small $\mathcal{E}_i$}
    \end{overpic}\label{fig:workspace_cost_function_fig}
  }
  \hspace{-0.5cm}
  \subfloat[\footnotesize approximated signed distance functions]{
    \begin{overpic}[width=0.24 \textwidth]{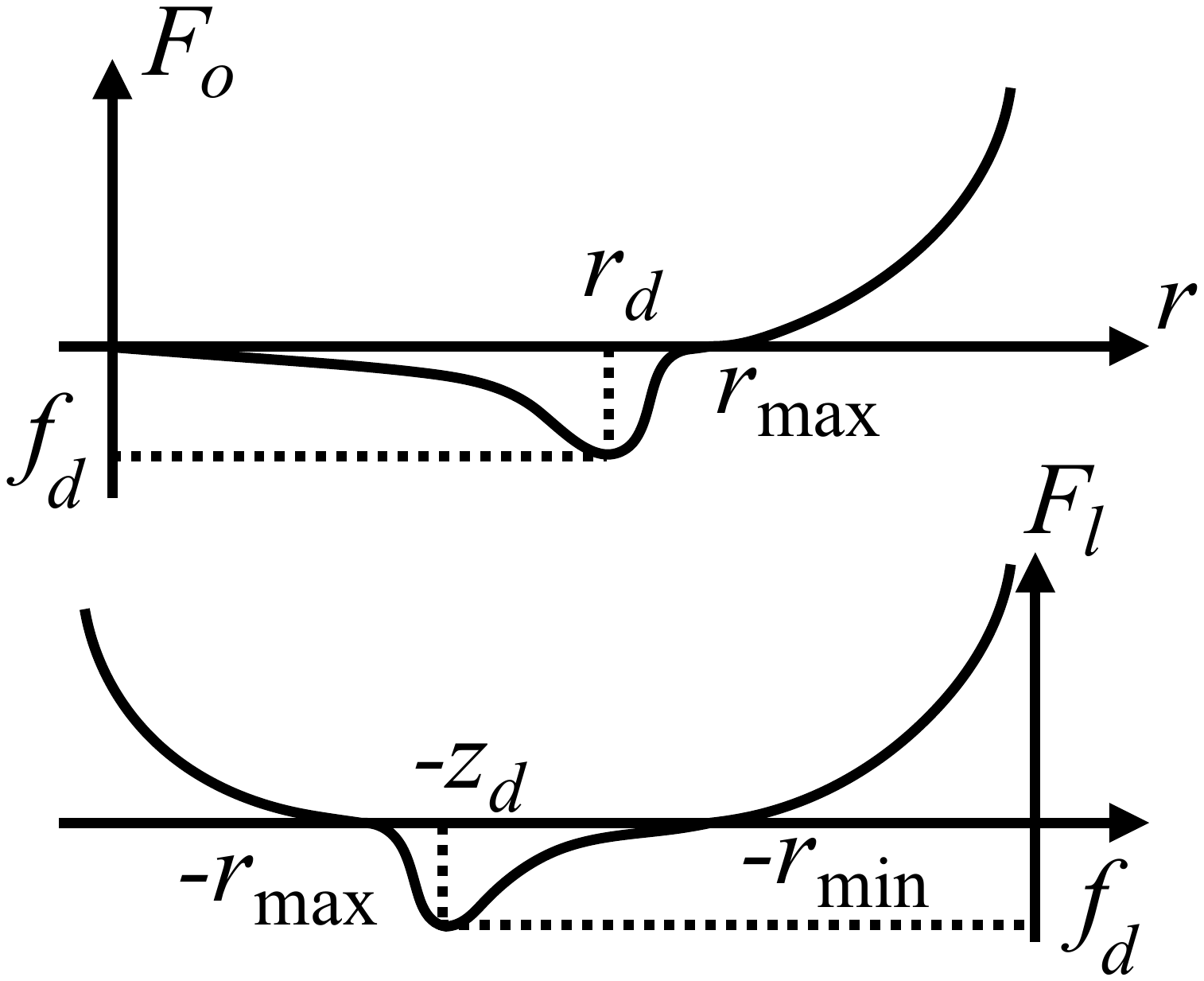}
      \put(82, 16){$\mathcal{E}_{i,z}$}
    \end{overpic}\label{fig:workspace_cost_function}
  }
  \caption{Illustration of two functions in the workspace cost.}\label{fig:illustration_of_sign_distance_function}
\end{figure}

As shown in Fig.\ \ref{fig:workspace_cost_function}, the specific form of $F_{o}$ and $F_{l}$ are as follows:
\begin{align}
  &F_{o}(\mathcal{Q}_{i}) =  \notag \\
  &\begin{cases}
    b_{o, 1} r^2 + a_{o, 1} r^3,      & 0 \!\leq\! r \!\leq\! r_{d} \\
    b_{o, 2} {(r\!-\!r_{\max})}^2 + a_{o, 2} {(r\!-\!r_{\max})}^3,
                                      & r_{d} \!\leq\! r \!\leq\! r_{\max} \\
    {(r\!-\!r_{\max})}^2,    & r_{\max} \!\leq\! r
  \end{cases}\label{eq:circ_cost}
  \\
  &F_{l}(\mathcal{Q}_{i}) = \notag \\
  &\begin{cases}
    {(\mathcal{E}_{i,z}\!+\!r_{\max})}^{2},
                    & \mathcal{E}_{i,z} \!\leq\! -r_{\max} \\
    b_{l,1}{(\mathcal{E}_{i,z}\!+\!r_{\max})}^{2} + \\ 
    \qquad \qquad a_{l,1} {\left(\mathcal{E}_{i,z}\!+\!r_{\max}\right)}^{3},
                    & -r_{\max} \!\leq\! \mathcal{E}_{i,z} \!\leq\! - z_{d} \\
    b_{l,2}{(\mathcal{E}_{i,z}\!+\!r_{\min})}^{2} + \\ 
    \qquad \qquad a_{l,2} {\left(\mathcal{E}_{i,z}\!+\!r_{\min}\right)}^{3}, 
                    & -z_{d} \!\leq\! \mathcal{E}_{i,z} \!\leq\! -r_{\min} \\
    {(\mathcal{E}_{i,z}\!+\!r_{\min})}^{2},
                    & -r_{\min} \!\leq\! \mathcal{E}_{i,z}
    \end{cases}\label{eq:line_cost}
\end{align}
where $r_{\max}$ and $r_{\min}$ are the parameters of the convex region. $\mathcal{E}_{i,x}$, $\mathcal{E}_{i,y}$, and $\mathcal{E}_{i,z}$ are three components of the control point $\mathcal{E}_i$. $r = \sqrt{\mathcal{E}^\top_i \mathcal{E}_i}$ is the 2-norm of the vector $\mathcal{E}_i$. 
By guaranteeing that $F_o(\mathcal{Q}_i)$ and $F_l(\mathcal{Q}_i)$ are continuously differentiable, the coefficients $a_{o,j}$, $b_{o,j}$, $a_{l,j}$, and $b_{l,j}$ ($j=1,2$) can be derived by combining $r_d$, $z_d$ and $f_d$, while $r_d$ and $z_d$ are related by the goal state of the end-effector, and $f_d$ is a constant.

\subsection{Yaw Rate Cost}
\begin{figure}[htbp]
  \centering
  \includegraphics[width=0.27\textwidth]{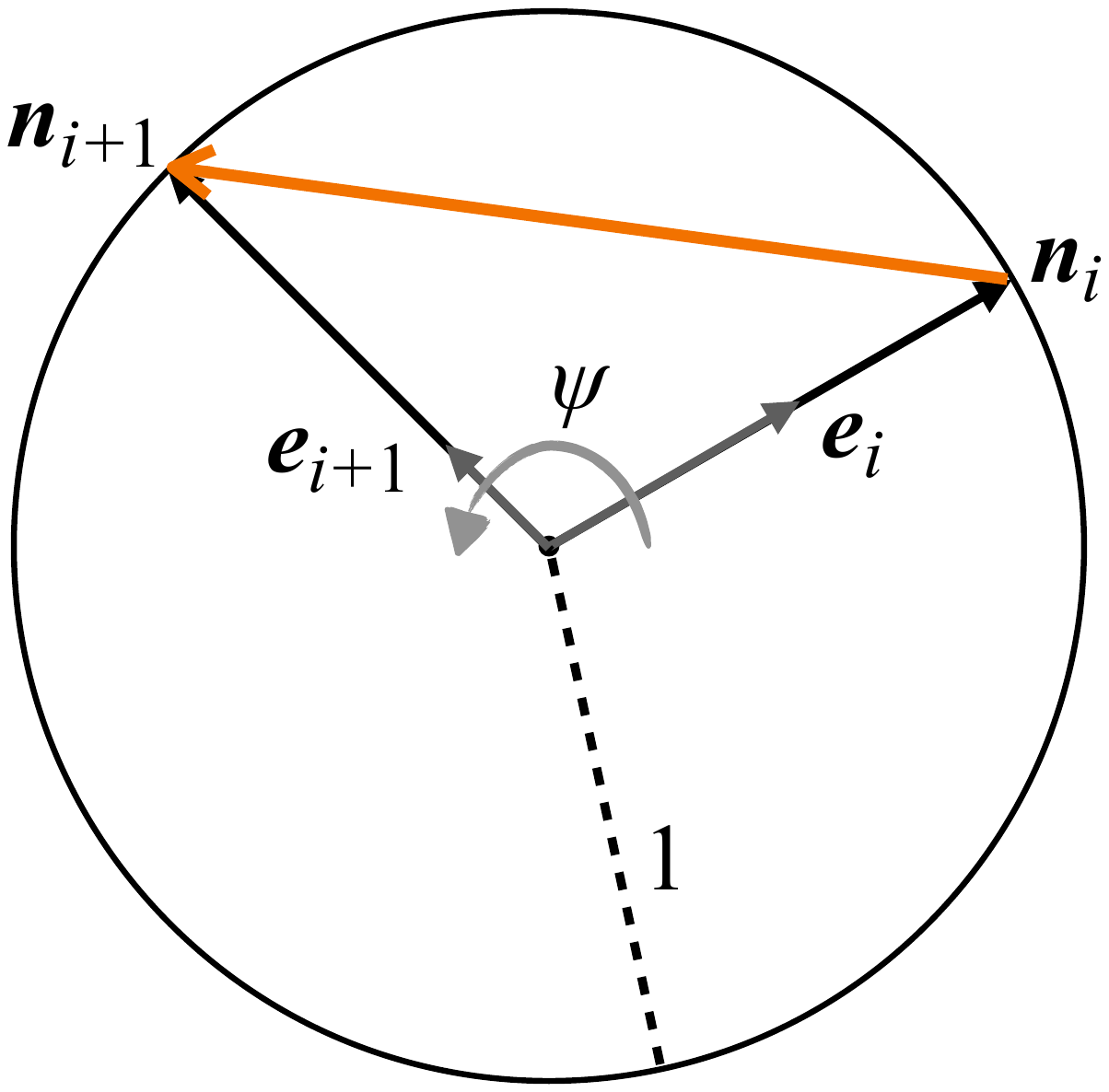}
  \caption{Illustration of the yaw rate cost. }\label{fig:yawrate_cost}
\end{figure}
As shown in Fig.\ \ref{fig:yawrate_cost}, $\bm e_i$ and $\bm e_{i+1}$ are the projections of the control points $\mathcal{E}_i$ and $\mathcal{E}_{i+1}$ on the $x$-$y$ plane, expressed as follows:
\begin{align*}
  \bm e_i = \begin{bmatrix}
    \mathcal{E}_{i,x} \\ \mathcal{E}_{i,y} \\ 0
  \end{bmatrix}, ~
  \bm e_{i+1} = \begin{bmatrix}
    \mathcal{E}_{i+1,x} \\ \mathcal{E}_{i+1,y} \\ 0
  \end{bmatrix}.
\end{align*}
$\bm n_i$ and $\bm n_{i+1}$ are the normalized vector of $\bm e_i$ and $\bm e_{i+1}$, respectively, which means that
\begin{align*}
  \bm n_i &= \frac{\bm e_i}{\|\bm e_i\|} = \begin{bmatrix}
      \frac{\mathcal{E}_{i,x}}{\sqrt{\mathcal{E}^2_{i,x} + \mathcal{E}^2_{i,y}}} \\
      \frac{\mathcal{E}_{i,y}}{\sqrt{\mathcal{E}^2_{i,x} + \mathcal{E}^2_{i,y}}} \\
      0
  \end{bmatrix}, \\
  \bm n_{i+1} &= \frac{\bm e_{i+1}}{\| \bm e_{i+1}\|} = \begin{bmatrix}
      \frac{\mathcal{E}_{i+1,x}}{\sqrt{\mathcal{E}^2_{i+1,x} + \mathcal{E}^2_{i+1,y}}} \\
      \frac{\mathcal{E}_{i+1,y}}{\sqrt{\mathcal{E}^2_{i+1,x} + \mathcal{E}^2_{i+1,y}}} \\
      0
  \end{bmatrix}.
\end{align*}
The yaw rate cost is formulated as
\begin{align}
    f_{y} = \sum_{i=s}^{N-s-1} F_{y}(\mathcal{Q}_i, \mathcal{Q}_{i+1})
          = \sum_{i=s}^{N-s-1} \| \bm n_{i+1} - \bm n_i \|^2.\label{eq:yawrate_cost}
\end{align}

\subsection{Smoothness Cost}
The smoothness cost is formulated as minimizing the normalization of the acceleration control points. The smoothness cost is formulated as
\begin{align}
  f_{s} = \sum_{i=s-2}^{N-s} F_s(\mathcal{Q}_i, \mathcal{Q}_{i+1}, \mathcal{Q}_{i+2})
        = \sum_{i=s-2}^{N-s} \| \mathcal{A}_i \|^2.\label{eq:smoothness_cost}
\end{align}
Different from the uniform B-spline, the acceleration control points $\mathcal{A}_i$ of the non-uniform B-spline is as follows:
\begin{align*}
  \mathcal{A}_i 
  = M_i \begin{bmatrix}
    \mathcal{E}_i \\ \mathcal{E}_{i+1} \\ \mathcal{E}_{i+2} 
  \end{bmatrix}
  = M_i \begin{bmatrix} 
    \mathcal{Q}_i \\ 
    \mathcal{Q}_{i+1} \\  
    \mathcal{Q}_{i+2}
  \end{bmatrix} - M_i \begin{bmatrix}
    \mathcal{X}_i \\ \mathcal{X}_{i+1} \\ \mathcal{X}_{i+2}
  \end{bmatrix}.
\end{align*}
The coefficients $M_i$ is defined as
\begin{align*}
  M_i &= [M_{i,0}, M_{i,1}, M_{i,2}] \\
  &= 
  \frac{s(s-1)}{t_{i+2,s-1}} \begin{bmatrix}
    \frac{1}{t_{i+1,s}} ~ -\!\left( \frac{1}{t_{i+2,s}} \!+\! \frac{1}{t_{i+1,s}} \right) ~ \frac{1}{t_{i+2,s}}
  \end{bmatrix},
\end{align*}
in which $t_{i,s}$ denotes the time span from $t_i$ to $t_{i+s}$, which means that $t_{i,s} = t_{i+s} - t_i$. The time knot vector $\mathcal{T} = [t_0, t_1, \ldots, t_{M}]{}^\top$ and all the control points $\mathcal{Q}_i$ have been derived from the multi-rotor's trajectory. 

\subsection{Obstacle Avoidance Cost}
The obstacle avoidance cost is formulated as the distance between the control point $\mathcal{Q}_i$ and the closet obstacle, formulated as follows: 
\begin{align}
    f_{d} &= \sum_{i=s}^{N-s} F_{d}(d(\mathcal{Q}_i)),
\end{align}
where $d(\mathcal{Q}_i)$ is the distance between the control point $\mathcal{Q}_i$ and the closest obstacle, which can be obtained by the ESDF map\cite{han2019iros}. The cost function on the $i$-th control point $F_{d}(d(\mathcal{Q}_i))$ is as follows:
\begin{align}
    F_{d}(d(\mathcal{Q}_i)) = \begin{cases}
        {( d(\mathcal{Q}_i) - d_{thr} )}^2, & d(\mathcal{Q}_i) \leq d_{thr} \\
        0, & d(\mathcal{Q}_i) > d_{thr}
    \end{cases}
\end{align}
where $d_{thr}$ is a distance threshold. 

The gradients of the above-mentioned cost functions with respect to the control points are derived in Appendix\ \ref{app:gradient}.


\section{Implementation and results}\label{sec:implementation_results}

In order to verify the effectiveness of the proposed method, the implementation details, simulation results and experimental results are introduced in this section.

\subsection{Algorithm Implementation and Experiment Setup}\label{subsec:algorithm_implementation_experiment_setup}
\subsubsection{Algorithm Implementation}
The order $s$ of the B-spline curve is set to 3 and the optimization problem presented in\ \eqref{eq:opt_problem} is solved by \text{NLopt} Library\footnote{https://nlopt.readthedocs.io/en/latest/}. The simulation platform is adapted from the Fast-Planner framework\cite{zhou2019robust}, incorporating the multi-rotor dynamics model, random map generator, and point cloud rendering module. All simulations are conducted on an Intel Core i7$-$13700 CPU and GeForce GTX 3070 Ti GPU.\ To ensure a fair comparison, all computations are conducted with the same aforementioned computation capability. In real world experiments, all the state estimation, mapping and motion planning modules run on an Intel Core i7$-$13700 CPU with 16GB RAM.\

\subsubsection{Aerial Manipulator}
\begin{figure}[htbp]
  \centering
  \includegraphics[width=0.4\textwidth]{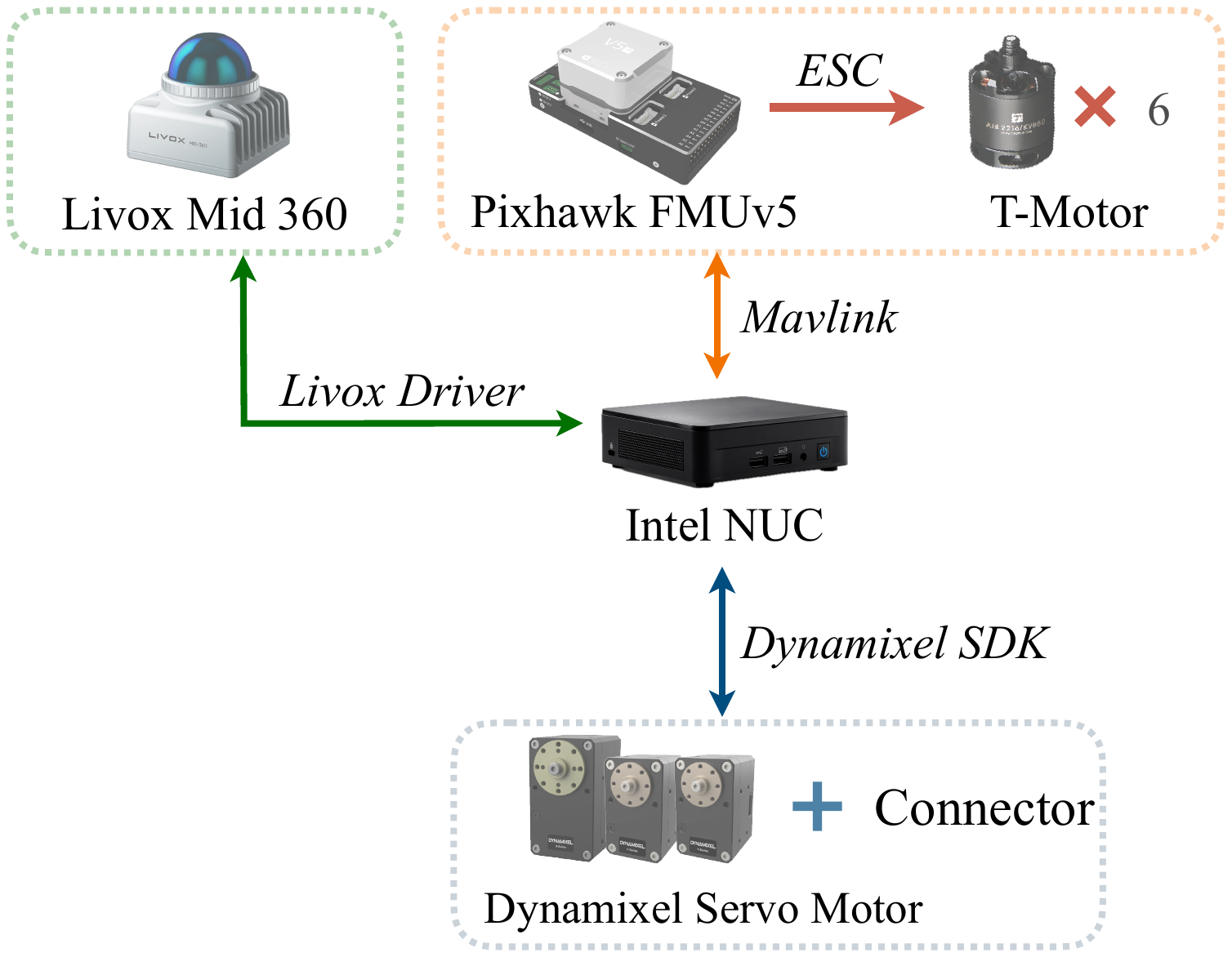}
  \caption{Experimental platform.}\label{fig:exp_platform}
\end{figure}
The aerial manipulator system consists of an F550 \text{hexrotor} and a custom-built robotic arm. Specifically, the \text{hexrotor} is equipped with a \text{Pixhawk} FMUv5 flight controller, which connects to the onboard computer via the Mavlink communication protocol. The flight controller employs the cascaded P-PID controller incorporating the feedforward term from the robotic arm, which is embedded within the \text{Pixhawk} FMUv5\footnote{https://ardupilot.org/copter/docs/common-cuav-v5plus-overview.html}. The flight controller powers six pairs of T-Motors and 10-inch propellers through the Electronic Speed Controllers (ESCs). The robotic arm is conducted by \text{Dynamixel} servo motors and a few self-designed connectors.

\subsubsection{State Estimation and Mapping}
The Livox Mid 360 LiDAR is mounted on the multi-rotor and integrated using the Livox Driver\footnote{https://github.com/Livox-SDK/Livox-SDK}. Fast-lio2\cite{xu2022tro} is employed to estimate the odometry of the multi-rotor and to generate a dense point cloud map. 
To enhance the motion planning algorithm, we increase the frequency of the multi-rotor's odometry estimation provided by Fast-lio2. The state estimation for the robotic arm is handled by the \text{Dynamixel} SDK, while the end-effector's position is computed using the robotic arm's forward kinematics.

\subsection{Simulation Results}\label{subsec:simulation_results}
\begin{figure}[htbp]
  \centering
  \includegraphics[width=0.42\textwidth, trim=0 0 0 0, clip]{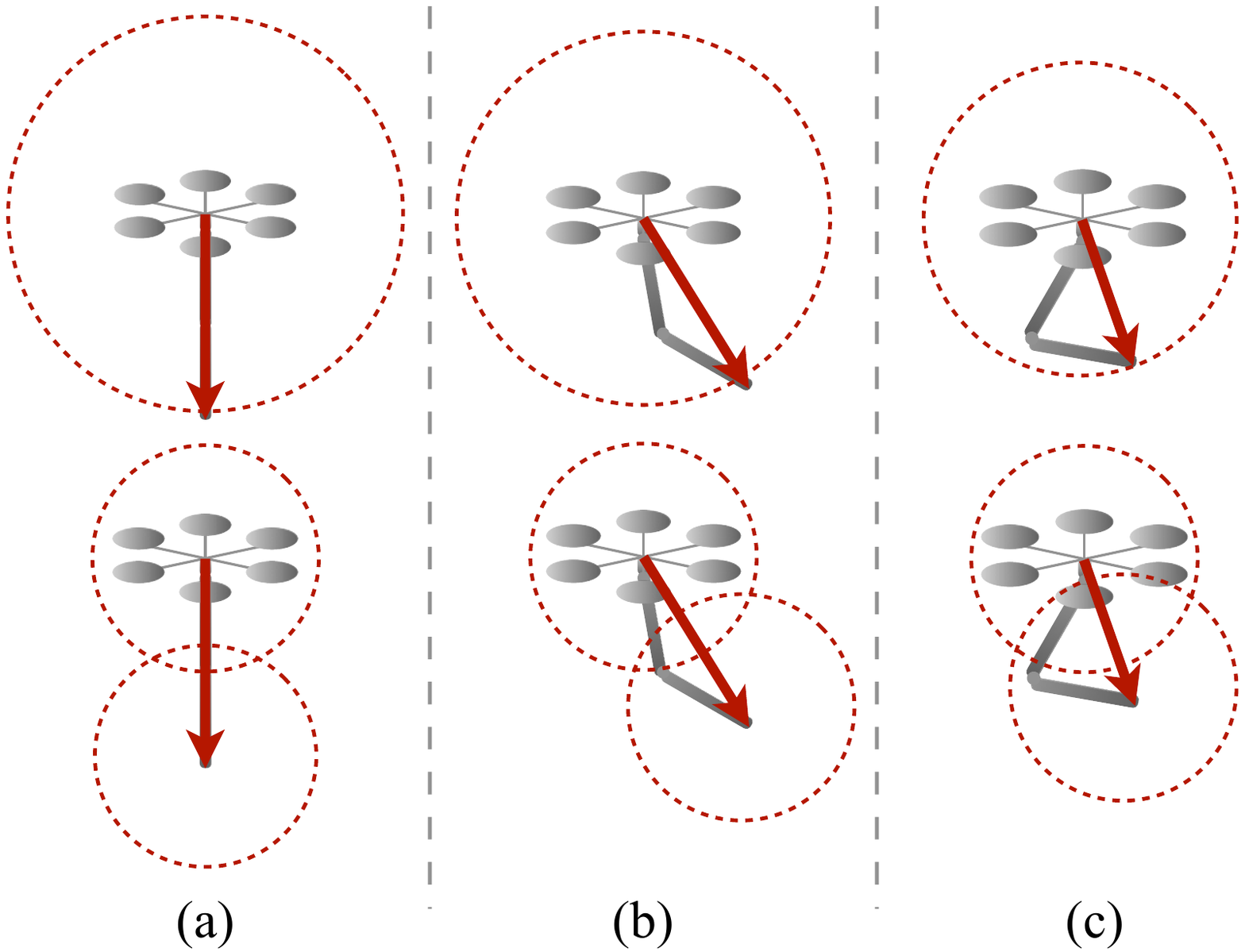}
  \caption{Illustration of the simulation scenario.}\label{fig:illustration_simulation_scenario}
\end{figure}
\begin{figure}[htbp]
  \centering
  \includegraphics[width=0.4\textwidth]{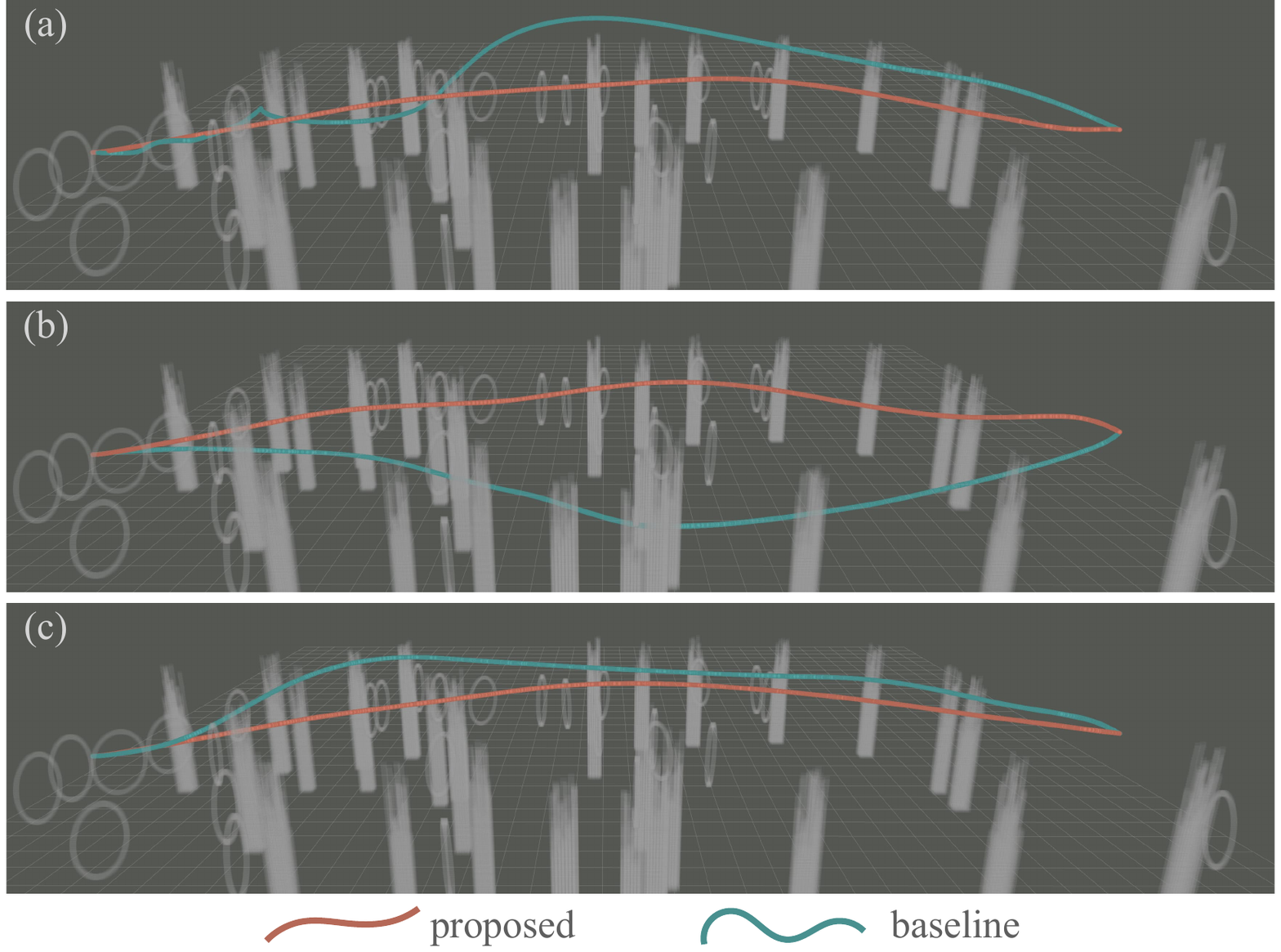}
  \caption{Multi-rotor's travel trajectories in different scenarios.}\label{fig:simulation_scenario_result}
\end{figure}
In this section, we evaluate the proposed method in three scenarios, as illustrated in Fig.\ \ref{fig:illustration_simulation_scenario}. Scenario (a), (b) and (c) depict three different initial and goal positions for the robotic arm. The red arrows extending from the center of the multi-rotor to the arm's end-effector represent the maximum distance from the multi-rotor's center to its outer edge. The proposed method incorporates trajectory planning for both the multi-rotor and the robotic arm, whereas the baseline method keeps the robotic arm fixed, which requires a larger inflation radius. Fig.\ \ref{fig:simulation_scenario_result} shows the multi-rotor's travel trajectories across the three scenarios within the same environment.

\begin{table}[h]
  \centering
  \caption{Quantitative Comparison}\label{tab:quantative_comparison}
  \renewcommand{\arraystretch}{1.2}
  \resizebox{\columnwidth}{!}{
    \begin{tabular}{cc|cc|ccc}
    \toprule[1pt] 
    \hline
    \multirow{2}{*}{Scenario} & \multirow{2}{*}{Method} 
    & \multicolumn{2}{c|}{Travel Trajectory}  & \multicolumn{3}{c}{Computation Time (ms)}\\
    \cline{3-7}
    &        &   Length (m)      &   Time (s) & multi-rotor & robotic arm & total \\
    \hline 
    \rowcolor{gray!20} \cellcolor{white!100} \multirow{2}{*}{(a)} 
    & proposed   & \textbf{42.52} & \textbf{15.79}  & 0.79   & 0.22  & \textbf{1.01} \\
    & baseline   & 49.63          & 22.87           & 1.73   & {-}   & 1.73          \\ 
    \hline 
    \rowcolor{gray!20} \cellcolor{white!100} \multirow{2}{*}{(b)}
    & proposed   & \textbf{42.58} & \textbf{16.03}  & 1.00   & 0.21  & \textbf{1.21} \\
    & baseline   & 44.08          & 17.72           & 1.55   & {-}   & 1.55          \\
    \hline 
    \rowcolor{gray!20} \cellcolor{white!100} \multirow{2}{*}{(c)}
    & proposed   & \textbf{42.39} & \textbf{15.88}  & 1.05   & 0.12  & \textbf{1.17} \\
    & baseline   & 43.82          & 16.65           & 1.22   & {-}   & 1.22          \\
    \hline \bottomrule[1pt]
    \end{tabular}
  }
\end{table}
The quantitative comparison results are presented in TABLE\ \ref{tab:quantative_comparison}. From the above comparison, the proposed method achieves shorter trajectory length and flight time for the multi-rotor's trajectory than the baseline method, which is intuitive. 
In the baseline method, the robotic arm is assumed to remain in a fixed configuration throughout the entire trajectory. To account for potential collisions without modeling the manipulator's motion, the entire aerial manipulator is conservatively enclosed within a large bounding sphere.
The quantitative results also show that the baseline method exhibits shorter travel distance and time in scenarios (b) and (c), attributed to the reduced inflation radius in these cases. It's more like an ablation study. 

In addition to travel trajectories, computation time is compared as well. The computation time in TABLE\ \ref{tab:quantative_comparison}\ is the average time of each planning process in the whole travel trajectory. The baseline method does not include computation time for the robotic arm, while the proposed method includes both the multi-rotor and the robotic arm's computation time.
The proposed method has a shorter time than the baseline method in all scenarios. This is because the baseline method requires trajectory planning for the multi-rotor in a more constrained environment, leading to higher computation cost in both path searching and trajectory optimization. 

\begin{figure}[htbp]
  \centering
  \includegraphics[width=0.4\textwidth, trim=0 0 0 1.0cm, clip]{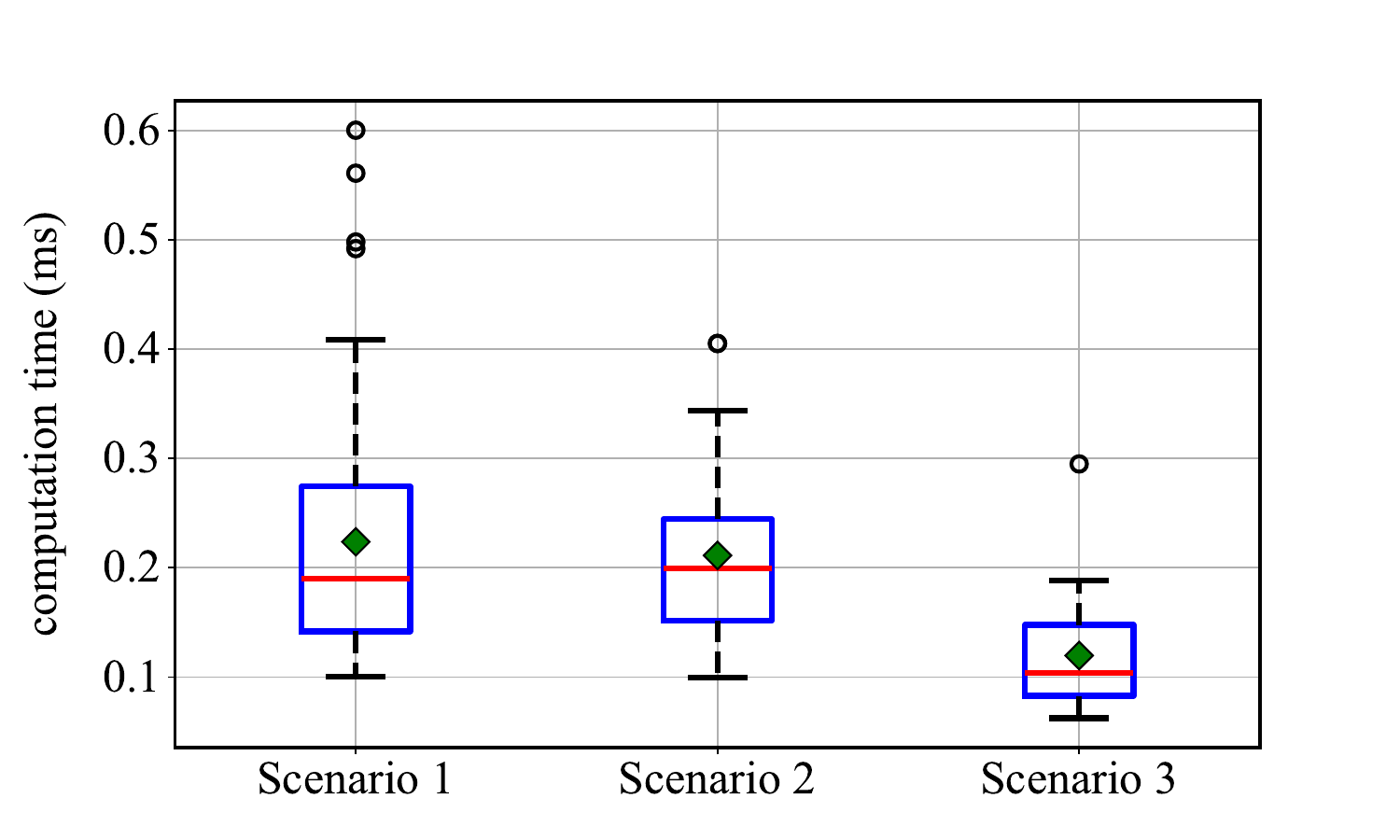}
  \caption{The detail of the robotic arm's computation time.}\label{fig:boxplot}
\end{figure}

For the purpose of further exploring the impact of the proposed method on the computation time of the robotic arm, we present a detailed boxplot in Fig.\ \ref{fig:boxplot}. The boxplot provides a detailed illustration of the computation time of Algorithm\ \ref{alg:traj_opt} across different scenarios. Even in the most extreme case for scenario (a), the time remains \textit{less than 0.6 ms}.
\subsection{Experimental Results}\label{subsec:experimental_results}
\begin{figure}[htbp]
  \centering
  \includegraphics[width=0.42\textwidth, trim=0 0 0 6cm, clip]{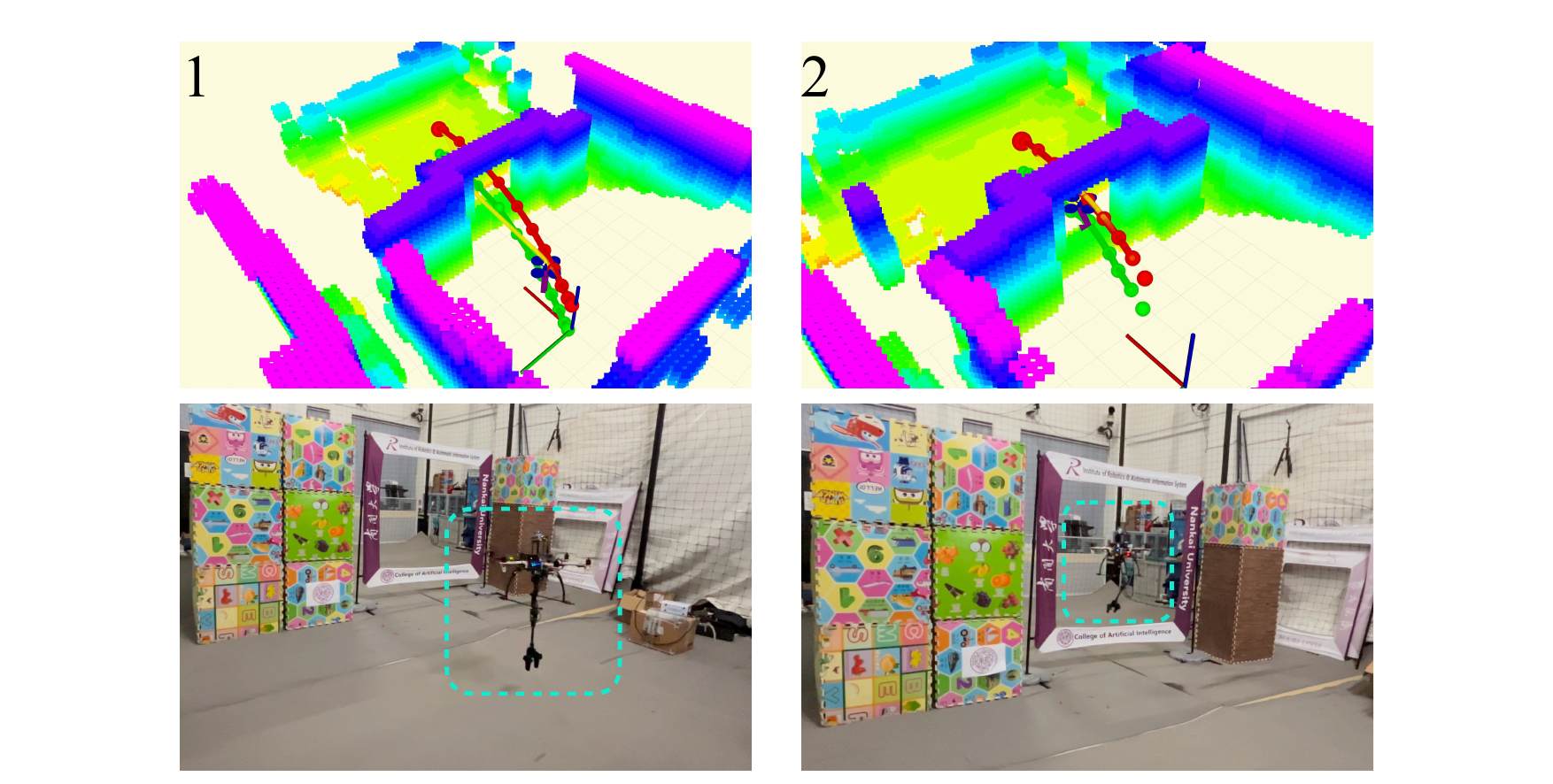}
  \caption{The snapshots and visualization of Experiment 1.}\label{fig:exp1_snapshot}
\end{figure}

\begin{figure}[htbp]
  \centering
  \includegraphics[width=0.488\textwidth, trim=0 0 0 7cm, clip]{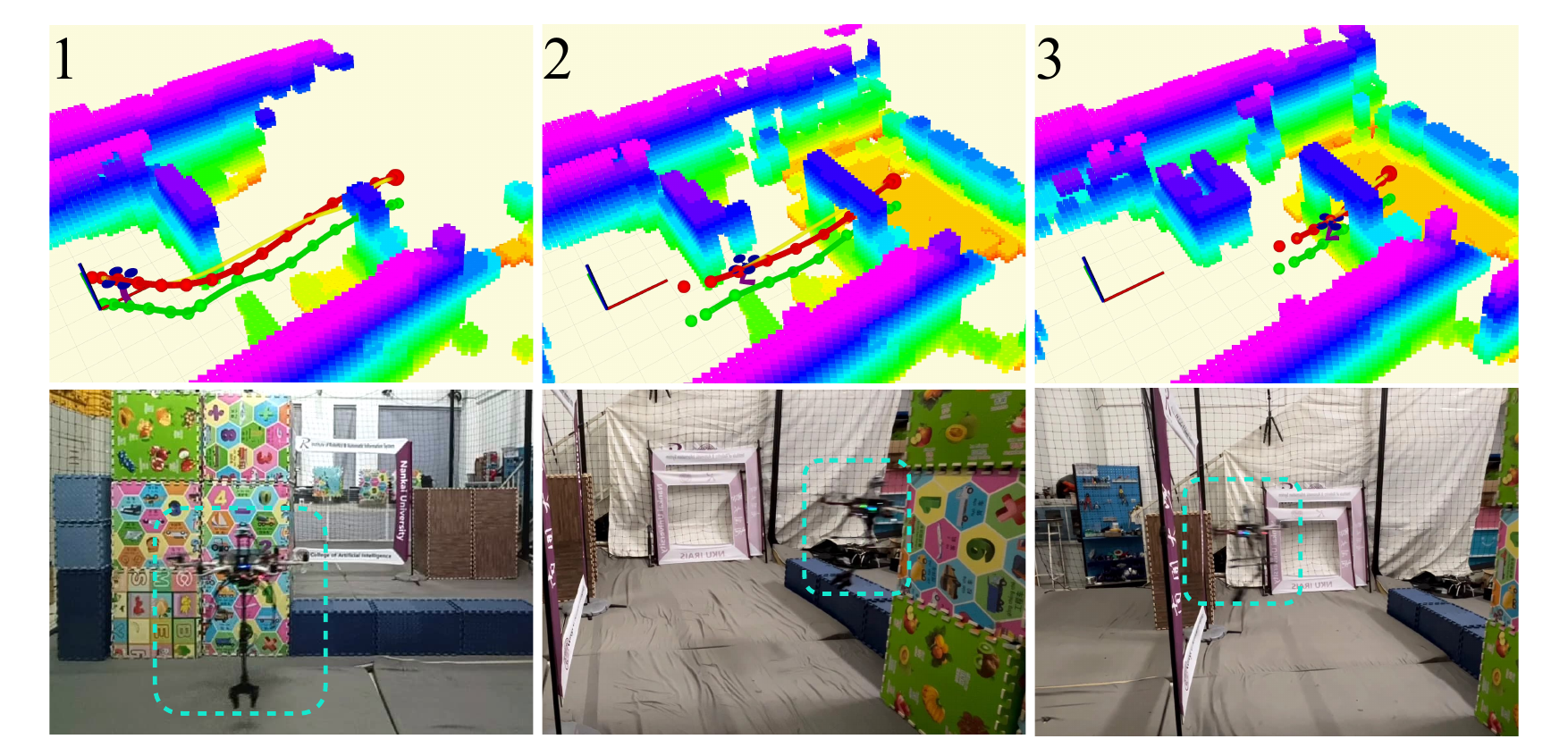}
  \caption{The snapshots and visualization of Experiment 2.}\label{fig:exp2_snapshot}
\end{figure}

In this paper, we present two fully autonomous flight experiments conducted in unknown environments, as illustrated in Fig.\ \ref{fig:exp1_snapshot} and Fig.\ \ref{fig:exp2_snapshot}. 

In Experiment 1, the aerial manipulator can successfully plan a collision-free trajectory for both the multi-rotor and the end-effector through a ring-shaped obstacle. 
In Experiment 2, the aerial manipulator starts behind the vertical obstacles, and it can successfully plan a collision-free trajectory that involves hurdling over the horizontal obstacles and navigating through an unknown ring-shaped obstacle.

To the best of our knowledge, this is the first instance of an aerial manipulator achieving autonomous flight in unknown environments. The experimental results demonstrate the effectiveness of the proposed method in real-world applications.



\section{Conclusion}\label{sec:conclusion}
In this paper, the motion planning problem for aerial manipulators is formulated as the positions of two points in the space, including the positions of the multi-rotor and the end-effector. By utilizing the planned and parameterized trajectory of the multi-rotor, we propose a real-time algorithm to plan the trajectory for the end-effector. An initial trajectory for the end-effector is generated by a second-order Bézier curve. Then, the gradient-based optimization method is used to optimize the trajectory of the end-effector. The linear and convex-hull properties of the B-spline curve ensure that the trajectory is smooth, collision-free and workspace-compatible. The simulation and experimental results show that the proposed method's effectiveness. Compared to motion planning strategies that only consider the multi-rotor, our approach yields less conservatism.

Although it is the pitch-pitch 2-DOF robotic arm that this paper focuses on, the proposed method can be easily extended to robotic arms with other configurations. In future work, we will develop a planner that simultaneously considers the positions of both the multi-rotor and the end-effector, and validate the algorithm in field environments.

\appendices\
\section{The gradient of the cost function}\label{app:gradient}
\subsection{Workspace cost}\label{app:gradient_workspace}
The gradient of the distance $r$ to the control point $\mathcal{E}_i$ is calculated as
\begin{align*}
    \frac{\partial r^2}{\partial r} \cdot \frac{\partial r}{\partial \mathcal{E}_i} = 2 r \cdot \frac{\partial r}{\partial \mathcal{E}_i} = 2 \mathcal{E}_i \quad 
    \Rightarrow \quad \frac{\partial r}{\partial \mathcal{E}_i} = \frac{\mathcal{E}_i}{r}.
\end{align*}
The gradient of the workspace cost $F_{w}(\mathcal{Q}_i)$ is given as follows:
\begin{align*}
    \frac{\partial f_{w}}{\partial \mathcal{Q}_i}
    ={}& \frac{h_{o} e^{h_{o} k F_{o}(\mathcal{Q}_i)}}{e^{h_o k F_o(\mathcal{Q}_i)} + e^{h_l k F_l(\mathcal{Q}_i)}} \frac{\partial F_{o}(\mathcal{Q}_i)}{\partial \mathcal{Q}_i}  \\
    &+ \frac{h_{l} e^{h_{l} k F_{l}(\mathcal{Q}_i)}}{e^{h_o k F_o(\mathcal{Q}_i)} + e^{h_l k F_l(\mathcal{Q}_i)}} \frac{\partial F_{l}(\mathcal{Q}_i)}{\partial \mathcal{Q}_i}.
\end{align*}
The gradient of the circle cost $F_{o}(\mathcal{Q}_i)$ is calculated as
\begin{align*}
    &\frac{\partial F_{o}(\mathcal{Q}_i)}{\partial \mathcal{Q}_i}
    = \frac{\partial F_{o}(\mathcal{Q}_i)}{\partial r} \frac{\partial r}{\partial \mathcal{E}_i} \frac{\partial \mathcal{E}_i}{\partial \mathcal{Q}_i}
    = \frac{\partial F_{o}(\mathcal{Q}_i)}{\partial r} \frac{\mathcal{E}_i}{r}. \\
    &\frac{\partial F_{o}(\mathcal{Q}_{i})}{\partial r} =  \notag \\
    &\begin{cases}
        2 b_{o, 1} r + 3 a_{o, 1} r^2,      & 0 \leq r \leq r_{d} \\
        2 b_{o, 2} {(r\!-\!r_{\max})} + 3 a_{o, 2} {(r\!-\!r_{\max})}^2,
                                            & r_{d} \leq r \leq r_{\max} \\
        2 {(r\!-\!r_{\max})}.       & r_{\max} \leq r
    \end{cases}\label{eq:circ_cost_gradient}
\end{align*}
The gradient of the line cost $F_{l}(\mathcal{Q}_i)$ is calculated as
\begin{align*}
    &\frac{\partial F_l(\mathcal{Q}_i)}{\partial\mathcal{Q}_i} 
    = \frac{\partial F_l(\mathcal{Q}_i)}{\partial\mathcal{E}_i} \frac{\partial \mathcal{E}_i}{\partial \mathcal{Q}_i}
    = \begin{bmatrix}
        0 \\  0 \\  \frac{\partial F_{l}(\mathcal{Q}_i)}{\partial \mathcal{E}_{i,z}}
    \end{bmatrix}. \\
    &\frac{\partial F_{l}(\mathcal{Q}_{i})}{\mathcal{E}_{i,z}} = \notag \\
    &\begin{cases}
    2 {(\mathcal{E}_{i,z}\!+\!r_{\max})}, & -r_{\max} \leq \mathcal{E}_{i,z}, \\
    2 b_{l,1}{(\mathcal{E}_{i,z}\!+\!r_{\max})} + \\ 
    \qquad \qquad 3 a_{l,1} {\left(\mathcal{E}_{i,z}\!+\!r_{\max}\right)}^{2},  & -r_{\max} \leq \mathcal{E}_{i,z}\leq - z_{d} \\
    2 b_{l,2}{(\mathcal{E}_{i,z}\!+\!r_{\min})} + \\ 
    \qquad \qquad 3 a_{l,2} {\left(\mathcal{Q}_{i,z}\!+\!r_{\min}\right)}^{2},  & -z_{d} \leq \mathcal{E}_{i,z}\leq - r_{\min} \\
    2 {(\mathcal{E}_{i,z}\!+\!r_{\min})}. & -r_{\min} \leq \mathcal{E}_{i,z}
    \end{cases}
\end{align*}

\subsection{Yaw rate cost}\label{app:gradient_yawrate}
The gradient of the yaw rate cost $F_{y}(\mathcal{Q}_i, \mathcal{Q}_{i+1})$ is given as follows:
\begin{align*}
  \frac{\partial F_y}{\partial \mathcal{Q}_{i}}
  &= -2 {\left[{\left( \bm n_{i+1} \!-\! \bm n_i \right)}^\top \frac{\partial \bm n_i}{\partial \mathcal{Q}_{i}}\right]}^\top \\
  &= -2 {\left[{\left( \bm n_{i+1} \!-\! \bm n_i \right)}^\top \frac{\partial \bm n_i}{\partial \mathcal{E}_{i}}\right]}^\top \\
  &= -2 {\left(\frac{\partial \bm n_i}{\partial \mathcal{E}_{i}}\right)}^\top \left( \bm n_{i+1} - \bm n_i \right). \\
  \frac{\partial F_y}{\partial \mathcal{Q}_{i+1}} 
  &= 2 {\left[{\left( \bm n_{i+1} \!-\! \bm n_i \right)}^\top \frac{\partial \bm n_{i+1}}{\partial \mathcal{Q}_{i+1}}\right]}^\top  \\
  &= 2 {\left[{\left( \bm n_{i+1} \!-\! \bm n_i \right)}^\top \frac{\partial \bm n_{i+1}}{\partial \mathcal{E}_{i+1}}\right]}^\top \\
  &= 2 {\left(\frac{\partial \bm n_{i+1}}{\partial \mathcal{E}_{i+1}}\right)}^\top \left( \bm n_{i+1} - \bm n_i \right).
\end{align*}
where, 
\begin{align*}
  \frac{\partial \bm n_i}{\partial \mathcal{E}_i} 
  &= \begin{bmatrix}
    \frac{\partial \bm n_{i,x}}{\partial \mathcal{E}_{i,x}} & \frac{\partial \bm n_{i,x}}{\partial \mathcal{E}_{i,y}} & 0 \\
    \frac{\partial \bm n_{i,y}}{\partial \mathcal{E}_{i,x}} & \frac{\partial \bm n_{i,y}}{\partial \mathcal{E}_{i,y}} & 0 \\
    0 & 0 & 0
  \end{bmatrix} \\
  &= \begin{bmatrix}
    \frac{\mathcal{E}^2_{i,y}} {{(\mathcal{E}^2_{i,x} + \mathcal{E}^2_{i,y})}^{\frac{3}{2}}} & \frac{- \mathcal{E}_{i,x} \mathcal{E}_{i,y}} {{(\mathcal{E}^2_{i,x} + \mathcal{E}^2_{i,y})}^{\frac{3}{2}}} & 0 \\
    \frac{- \mathcal{E}_{i,x} \mathcal{E}_{i,y}} {{(\mathcal{E}^2_{i,x} + \mathcal{E}^2_{i,y})}^{\frac{3}{2}}} & \frac{\mathcal{E}^2_{i,x}} {{(\mathcal{E}^2_{i,x} + \mathcal{E}^2_{i,y})}^{\frac{3}{2}}} & 0 \\
    0 & 0 & 0
  \end{bmatrix} = {\left(\frac{\partial \bm n_{i}}{\partial \mathcal{E}_{i}}\right)}^\top.
\end{align*}
It can be concluded that 
\begin{align*}
  \frac{\partial f_y}{\partial \mathcal{Q}_i} 
  = -&2 {\left( \frac{\partial \bm n_i}{\partial \mathcal{E}_i} \right)}^\top \left( \bm n_{i+1} - \bm n_i \right) \\ 
  + &2 {\left( \frac{\partial \bm n_{i}}{\partial \mathcal{E}_{i}} \right)}^\top \left( \bm n_i - \bm n_{i-1} \right).
\end{align*}

\subsection{Smoothness cost}\label{app:gradient_smoothness}
The gradient of the smoothness cost is calculated as
\begin{align*}
    \frac{\partial f_{s}}{\partial \mathcal{Q}_i}
    ={}& {\left(2 \mathcal{A}_i^\top \frac{\partial \mathcal{A}_i}{\partial \mathcal{E}_i} \frac{\partial \mathcal{E}_i}{\partial \mathcal{Q}_i}\right)}^\top
    + {\left(2 \mathcal{A}_i^\top \frac{\partial \mathcal{A}_{i-1}}{\partial \mathcal{E}_i} \frac{\partial \mathcal{E}_i}{\partial \mathcal{Q}_i}\right)}^\top \\
    &+ {\left(2 \mathcal{A}_i^\top \frac{\partial \mathcal{A}_{i-2}}{\partial \mathcal{E}_i} \frac{\partial \mathcal{E}_i}{\partial \mathcal{Q}_i}\right)}^\top
    \\
    ={}& 2 {\left(\frac{\partial \mathcal{A}_i}{\partial \mathcal{E}_i}\right)}^\top \mathcal{A}_i 
    + 2 {\left(\frac{\partial \mathcal{A}_{i-1}}{\partial \mathcal{E}_i}\right)}^\top \mathcal{A}_{i-1} \\
    &+ 2 {\left(\frac{\partial \mathcal{A}_{i-2}}{\partial \mathcal{E}_i}\right)}^\top \mathcal{A}_{i-2}.
    \\
    ={}& 2 M_{i,0} \mathcal{A}_i + 2 M_{i-1,1} \mathcal{A}_{i-1} + 2 M_{i-2,2} \mathcal{A}_{i-2}.
\end{align*}

\subsection{Obstacle Avoidance cost}\label{app:gradient_obstacle}
The gradient of the obstacle avoidance cost $F_{o}(\mathcal{Q}_i)$ is calculated as
\begin{align*}
    \frac{f_d}{\partial \mathcal{Q}_i} 
    = 2 \left(d(\mathcal{Q}_i) - d_{\text{thr}}\right) \frac{\partial d(\mathcal{Q}_i)}{\partial \mathcal{Q}_i},
\end{align*}
where the gradient of the distance $d(\mathcal{Q}_i)$ to the control point $\mathcal{Q}_i$ can be obtained in the ESDF map\cite{han2019iros} directly.

\bibliographystyle{ieeetr}
\bibliography{ref/motionplan_uav, ref/motionplan_uam, ref/else}

\end{document}